\documentclass[preprint,12pt]{elsarticle}

\usepackage{multicol}%
\usepackage[utf8]{inputenc} 
\usepackage{amssymb}
\usepackage{graphicx}
\usepackage{float}
\restylefloat{table}
\usepackage{placeins}
\usepackage{float}
\usepackage{wrapfig}
\usepackage{algorithm}
\usepackage{algorithmic}
\usepackage{array}
\usepackage{color}
\usepackage{csquotes}
\usepackage{capt-of}
\usepackage{longtable}
\usepackage{caption}
\usepackage{amsmath}
\usepackage{tikz}
\usepackage[utf8]{inputenc}
\usepackage{float}
 \usepackage[strict]{changepage}
\usepackage[skip=0pt]{caption}
\usetikzlibrary{positioning,calc}
\usetikzlibrary{shapes,fit} 
\usetikzlibrary{shapes,arrows}
\usepackage{graphicx,subfigure}

\usepackage{graphicx}
\usepackage{color}
\usepackage{mathtools}
\definecolor{lightgray}{gray}{0.5}
\makeatletter
\newcommand{\removelatexerror}{\let\@latex@error\@gobble}
\makeatother
\newcolumntype{C}{>{\centering\arraybackslash}p{5em}}

\newcommand{\beq}{\begin{equation}}
\newcommand{\eeq}{\end{equation}}
\newcommand{\bea}{\begin{algorithmic}}
\newcommand{\eea}{\end{algorithmic}}
\newcommand{\M}[1]{\boldsymbol{#1}}  
\newcommand{\V}[1]{\boldsymbol{#1}}

\renewcommand{\vec}{\mbox{vec}}

\usepackage{xparse}
\NewDocumentCommand{\ceil}{s O{} m}{%
  \IfBooleanTF{#1} 
    {\left\lceil#3\right\rceil} 
    {#2\lceil#3#2\rceil} 
}
\usepackage{cleveref}
\crefname{section}{§}{§§}
\Crefname{section}{§}{§§}

\begin{document}
\begin{frontmatter}
\title{Robust Regression For Image Binarization Under Heavy Noises and Nonuniform Background}


\author[ime]{Garret D. Vo}
\author[ime]{Chiwoo Park\corref{cor1}}
\cortext[cor1]{Corresponding author}
\ead{cpark5@fsu.edu, Tel. +1-850-410-6457}
\address[ime]{Department of Industrial and Manufacturing Engineering, Florida State University, 2525 Pottsdamer St. Tallahassee, FL 32310}

\begin{abstract}
This paper presents a robust regression approach for image binarization under significant background variations and observation noises. The work is motivated by the need of identifying foreground regions in noisy microscopic image or degraded document images, where significant background variation and severe noise make an image binarization challenging. The proposed method first estimates the background of an input image, subtracts the estimated background from the input image, and apply a global thresholding to the subtracted outcome for achieving a binary image of foregrounds. A robust regression approach was proposed to estimate the background intensity surface with minimal effects of foreground intensities and noises, and a global threshold selector was proposed on the basis of a model selection criterion in a sparse regression. The proposed approach was validated using 26 test images and the corresponding ground truths, and the outcomes of the proposed work were compared with those from nine existing image binarization methods. The approach was also combined with three state-of-the-art morphological segmentation methods to show how the proposed approach can improve their image segmentation outcomes.
\end{abstract}

\begin{keyword}
Image Binarization \sep Background Subtraction \sep Robust Regression \sep Document Image Analysis \sep Microscopy Image Analysis
\end{keyword}

\end{frontmatter}
\section{Introduction} \label{intro}
Image binarization is a problem of estimating the binary silhouette of foreground objects in a noisy image. A good image binarization solution has been useful in many different contexts such as optical character recognition \cite{lu2010document}, image segmentation using binarization \cite{otsu1979threshold}, and preliminary image segmentation prior to separating overlaps of foregrounds \cite{park2013segmentation}. In particular, many existing morphological image segmentation algorithms  \cite{park2013segmentation, schmitt2009morphological, zafari2015segmentation} require the binary silhouette as inputs to find markers to individual foreground objects. The quality of their image segmentation is significantly affected by the accuracy of the binary silhouette input, so achieving a good binary silhouette estimate is essential. However, image binarization can be challenging under real-life situations such as uneven background and noises. This paper is concerned with an image binarization problem under uneven background and high image noises. \\

\begin{figure}
\centering
  \subfigure[Microscope image with uneven illumination and medium noises]{    
	\includegraphics[height=0.34\textwidth]{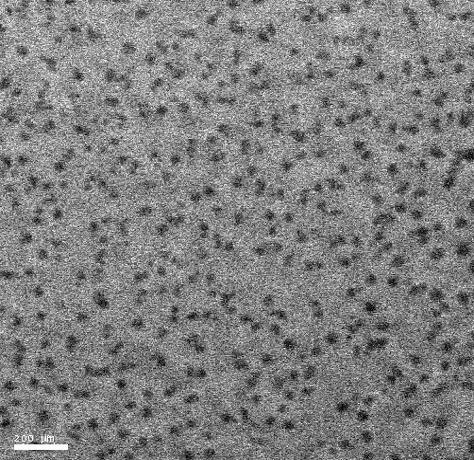}
	}	
  \subfigure[Microscope image with background tranisition and mild noises]{    
		\includegraphics[height=0.34\textwidth]{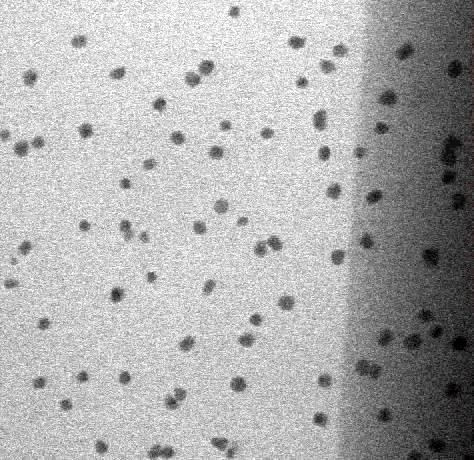}	
	}
	\subfigure[OCR document image with uneven illumination and mild noises]{    
			\includegraphics[height=0.3\textwidth]{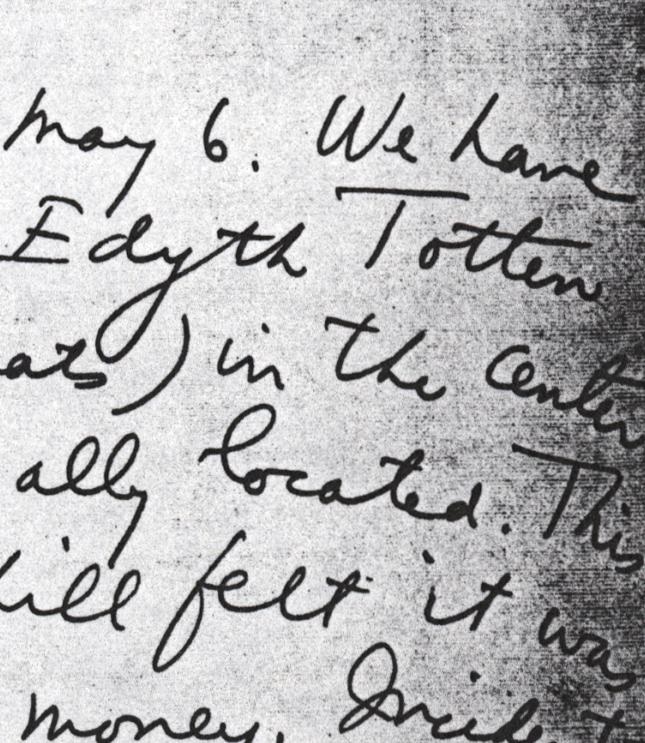}	
		}
			\subfigure[Handwritten text document image with document degradation]{    
					\includegraphics[height=0.3\textwidth]{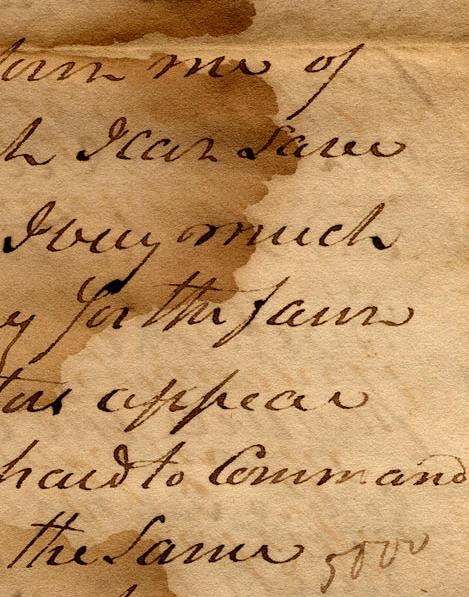}	
				}
\caption{Example Images with Nonuniform Background}
\label{fig0}
\end{figure}

Our first motivating example is to detect nanoparticles in heavily noisy microscope images. A typical electron microscopic image has non-uniform background intensities due to the spatial variation of electron beam radiations (e.g. Fig. \ref{fig0}-(a)) or due to the transitions between different background materials (e.g. Fig. \ref{fig0}-(b)). In addition, microscope images typically contain significant noises. A simple and popular approach for detecting nanoparticles is morphological segmentation \cite{park2013segmentation, schmitt2009morphological, zafari2015segmentation, malpica1997applying, tek2005blood, adiga2001efficient}. The approach applies a series of morphological image operations to roughly locate or mark overlapping foreground objects in an input image, which guides the subsequent image segmentation based on watershed or other sophisticated methods. The first marking step takes the binary silhouette of foreground regions as an input, or the binary silhouette is internally generated by their built-in image thresholding units, which did not work very well under a varying image background and a low signal-to-noise ratio. With a good method for estimating the binary silhouette, the accuracy of many morphological image segmentation methods can be greatly improved. \\

The second motivating example is to detect typed texts or handwritten texts in document images. The image binarization is an important preliminary stage for document image analysis. Typical text document images have nonuniform background due to degradation of text documents as shown in Fig. \ref{fig0}-(c) and -(d). An effective solution of an image binarization problem under nonuniform background can be very useful to extract texts from document images. \\

For binarizing an image with varying background and significant noise,  we propose a two-stage approach. The first stage is to estimate the varying background of an input image, which is then subtracted from the input image. With successful background estimation, the subtraction result would have a flat background. The second stage is to select a global threshold to threshold the subtraction result to an binary image. We propose robust methods to estimate the background and the global threshold.\\

The remainder of this paper is organized as follows. Section \ref{sec:review} reviews the related research and discusses the contribution of the proposed approach. Section \ref{sec:method} presents our approach. Section \ref{sec:app} presents the numerical performance of the proposed approach using 26 test images, with comparison to nine state-of-the-art methods. Section \ref{con} concludes this paper with discussion.

\section{Related Research} \label{sec:review}
The image thresholding or image binarization problem has been extensively studied in computer vision. The simplest approach selects and applies one \cite{otsu1979threshold} or two thresholds \cite{lai2014efficient} for an entire input image to classify individual pixels into foreground and background pixels. This is called the global thresholding approach. Depending on how threshold(s) are selected, the approach can be categorized into histogram shape-based, clustering-based, entropy-based, object attribute-based and spatial methods; several comprehensive reviews can be found in literature \cite{sezgin2004survey, stathis2008evaluation}. The approach does not work well when an input image has background variations due to illumination effects or image degradation. \\

For images with nonuniform background, a local adaptive thresholding is more suitable. A popular approach for the local thresholding selects a threshold for each pixel, based on the local statistics within the neighborhood window centered at the pixel. Nikbalt \cite{niblack1985introduction} used the local mean $M$ and the local standard deviation $S$ to choose the local threshold $M + k \times S$, where $k$ is a user-defined constant. A weakness of this local approach is that if foreground objects are sparse in an input image, a lot of background noises would remain in the outcome binary image \cite{gatos2006adaptive}. To overcome this weakness, Sauvola \cite{sauvola2000adaptive} modified the threshold to $M \times (1+0.5(1-S/128))$. Phansalskar \cite{phansalkar2011adaptive} further modified it to deal with low contrast images. \\

Contrast threshold is also a popular local thresholding approach, which estimates and uses local image contrasts. The local image contrast of a pixel is often defined as the range of the intensities of the pixel's neighborhood. If the contrast is high, the pixel is likely to belong to a transition area in between foregrounds and backgrounds, and it belongs to foregrounds or backgrounds otherwise. Therefore, local image contrasts are often thresholded, and low contrast regions are classified into foregrounds or backgrounds depending on some local statistics \cite{bernse1986dynamic}. Su \cite{su2010binarization} first computed a local contrast image using the local minimum and maximum within a local neighborhood window and then thresholded the contrast image to identify high contrast pixels. High contrast pixels were regarded as boundary pixels of foregrounds, and their intensities were used to determine local thresholds. Su \cite{su2013robust} combined an image contrast map and an edge detector to detect high contrast edge pixels, and the mean and standard deviation of the edge pixel intensities were used to define local thresholds.\\

Background subtraction is another approach, which first estimates an image background intensity surface and then applies a global thresholding on the outcome of subtracting the background intensity surface from an input image. Gatos \cite{gatos2006adaptive} roughly split an input image into background and foreground pixels, and the background pixel intensities were interpolated to define a background intensity surface and the corresponding local threshold policy. The major weakness of this approach is that it is difficult to achieve a good initial separation of the background region when an input image has uneven background and heavy noises. Polynomial smoothing \cite{mieloch2005dynamic} is a background subtraction approach that does not require any rough estimates of foreground and background regions. Lu \cite{lu2010document} applied the polynomial smoothing to directly estimate the background surface without any rough estimates of foreground and background regions, which was combined with an edge detection algorithm to determine local thresholds. \\

Edge-level thresholding was also popularly used for a local adaptive thresholding. Parker \cite{parker1991, parker1993thresholding} first located the edge pixels of foregrounds using an existing edge detector and smoothly interpolated the edge pixels to build spatially varying thresholds. As a similar approach, Chen \cite{chen2008double} used the Canny's edge detector to extract edge pixels, and a region growing was applied with the edge pixels as seeds. Ramirez-Ortegon \cite{ramirez2010transition} proposed the generalization of an edge pixel to the concept of transition pixel. Transition pixels are first identified, and the dark regions surrounded by the transition pixels are identified as foregrounds.  \\

Among other notable approaches, there is an approach based on the Markov random field modeling of an input image, which regards the target binary image as a binary Markov random field that minimizes a cost function. Howe \cite{Howe2013} combined the Laplacian of an input image and edge detection results to define the cost function. This approach became the basis for the first winner of the 2016 hand written document binarization contest \cite{pratikakis2016icfhr2016}. \\

The aforementioned local adaptive thresholding approaches have been successfully applied for many applications and validated using many benchmark datasets such as multiple DIBCO datasets \cite{pratikakis2016icfhr2016}. Among the existing approaches, we follow and advance the background subtraction approach that first estimates the background of an input image and then applies a global thresholding on the outcome of subtracting the background estimate from the input image. The main contribution of this paper is to advance the background subtraction approach with a novel robust background estimation method and a global threshold selection approach. More details are summarized as follows:
\begin{itemize}
\item The major contribution is to develop a robust background intensity surface estimation method. The existing approaches estimate the background intensity surface by interpolating the intensities of background pixels \cite{gatos2006adaptive} or edge pixels \cite{parker1993thresholding}, so a prior identification of background pixels or edge pixels is required. However, the prior identification is challenging under uneven background and heavy noises. The global polynomial smoothing  \cite{lu2010document} is a method that does not require any prior identification of edges or backgrounds, but it still requires edge detections for the subsequent foreground detection. In addition, we found that the method generates some image artifacts on its background estimate, mainly because the smoothing was individually applied to each row and column of an input image. Our new approach provides a more robust option for background estimation. It borrows a robust regression concept to formulate and solve the background estimation, which is less affected by other outlying image features such as foreground pixels and noise pixels. In addition, the proposed approach does not require prior identification of edges or backgrounds.  
\item For the second thresholding step of our approach, we propose a global threshold selector distinct from many existing binarization approaches. Following the signal processing literatures \cite{donoho1994threshold}, we formulate an image thresholding problem as a sparse regression problem to recover the true signal from a noisy signal, where the choice of a threshold is related to the determination of the signal sparsity parameter. We use a model selection criterion for selecting the sparsity parameter and thus the corresponding threshold. This approach is distinct from the existing binarization approaches that use the histogram of image intensities or statistics of edge pixels. 
\item The practical values of the proposed approach is (a) to provide a robust option for image binarization for multiple contexts including document image binarization and microscopy analysis and (b) to improve the accuracy of the existing morphological image segmentation methods that require an accurate binary silhouette of foregrounds as inputs. We validated these points with 26 benchmark images, comparing to nine existing methods. 
\end{itemize}

\section{Method} \label{sec:method}
Let $Y(i,j)$ denote the $(i,j)$th pixel intensity of an input image of size $m \times n$, where foregrounds look darker than backgrounds. A local adaptive thresholding finds a local threshold $T(i,j)$ for the $(i,j)$th  pixel to threshold the input image into a binary image,
\begin{equation} \label{eq:localthresh}
B(i,j) = \begin{cases} 
1 & \mbox{ for } Y(i,j) \le T(i,j)\\
0 & \mbox{ for } Y(i,j) > T(i,j).
\end{cases}
\end{equation}
All image pixel $(i,j)$'s where $B(i,j)=1$ are foreground pixels, and the other pixels are background pixels. The background subtraction approach for a local thresholding  applies the following threshold \cite{gatos2006adaptive},
\begin{equation*}
T(i,j) = L(i, j) + \tau,
\end{equation*}
where $L(i, j)$ represents the background intensity at the $(i,j)$th pixel, and $\tau$ is a global threshold. When $L(i, j)$ is known, the original thresholding \eqref{eq:localthresh} is equivalent to applying the following global thresholding to the subtraction of background $L(i, j)$ from the input $Y(i,j)$, 
\begin{equation} \label{eq:localthresh2}
B(i,j) = \begin{cases} 
1 & \mbox{ for } Y(i,j) - L(i,j) \le \tau\\
0 & \mbox{ for } Y(i,j) - L(i,j) > \tau.
\end{cases}
\end{equation}
The methods of estimating $L(i,j)$ and $\tau$ determine the performance of the background subtraction approach. We propose novel approaches to estimate them. 
Section \ref{sec:bg_est} describes the estimation of $L(i,j)$, and Section \ref{sec:fgr} describes the estimation of $\tau$. 

\subsection{Robust regression for estimating $L(i,j)$} \label{sec:bg_est}
\begin{figure}
\centering
\includegraphics[width=\textwidth]{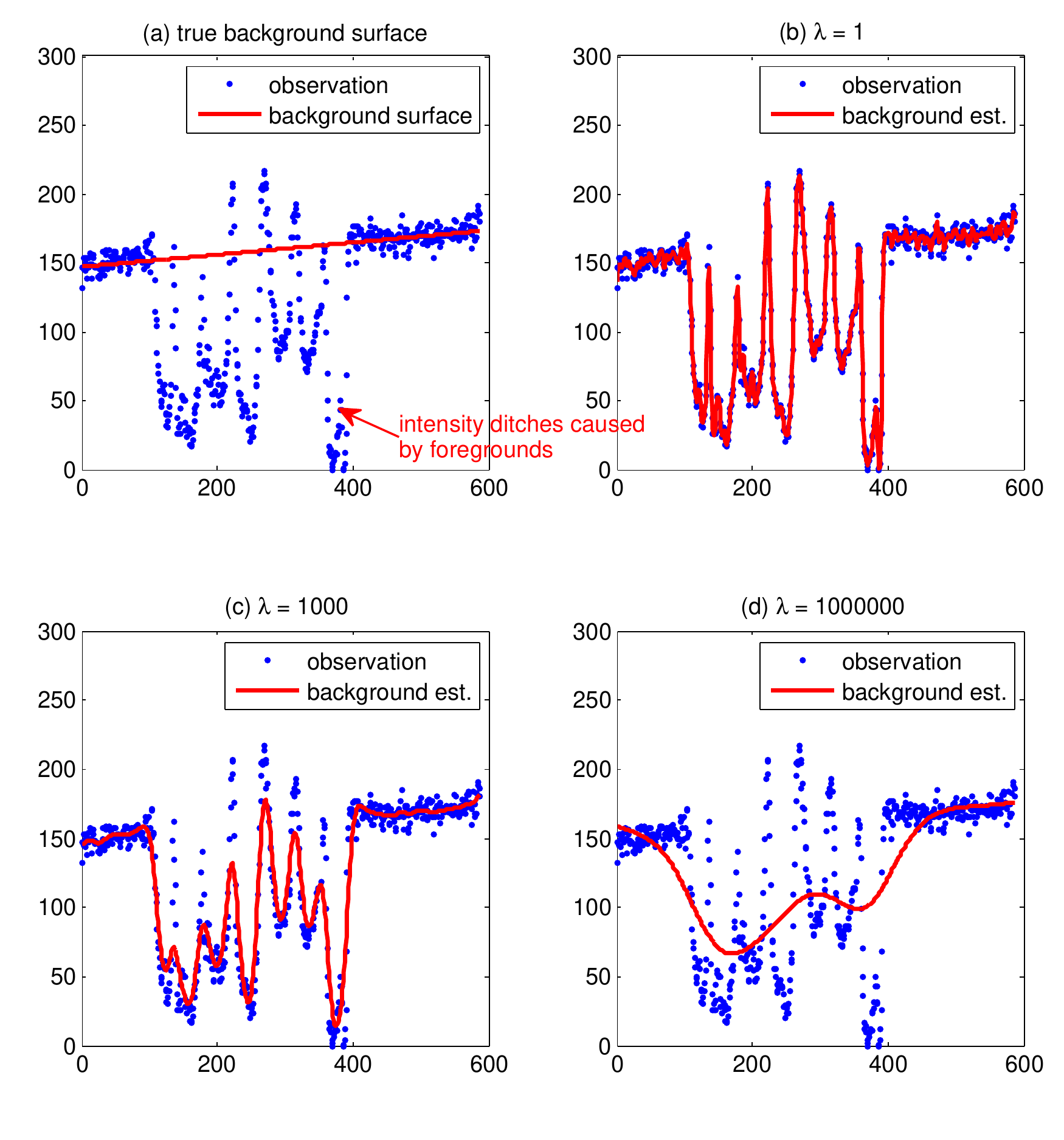}
\caption{Smoothing spline regression with square loss and different degrees of smoothness penalty $(\lambda)$.} \label{fig:smooth_spline}
\end{figure}

The input image $Y(i,j)$ is mixed with background $L(i, j)$, foreground $F(i,j)$ and noise $E(i,j)$ as
\begin{equation}
Y(i,j) = L(i,j) + F(i,j) + E(i,j),
\end{equation}
so estimating $L(i,j)$ hidden under foregrounds and noises is not straightforward. One possible approach is to roughly estimate the pixel locations where $F(i,j)\approx 0$ and interpolate the intensities of the pixels to estimate $L(i,j)$ like the existing approaches \cite{lu2010document,gatos2006adaptive}. However, finding the $(i,j)$'s that $F(i,j) \approx 0$ is as difficult as solving the original binarization problem \eqref{eq:localthresh}. Another possibility is to apply a smooth regression that interpolates $Y(i,j)$'s. In many applications, the background intensity surface $L(i,j)$ change smoothly over $(i, j)$, while $F(i,j)$ or $E(i,j)$ adds intensity jumps on the smooth background. Under the circumstances, estimating the smooth background can be possibly achieved by fitting a regression model to $Y(i,j)$'s with a square loss and a smoothness penalty, 
\begin{equation} \label{eq:smooth}
\min \sum_{i=1}^m \sum_{j=1}^n (Y(i,j) - L(i,j))^2 + \lambda \sum_{i=2}^{m-1} \sum_{j=2}^{n-1} |\nabla^2 L(i,j)|^2,
\end{equation} 
where $\nabla^2 L(i,j)$ is the $2$nd order derivative of $L$ at $(i,j)$, and $\lambda$ is a tuning parameter that determines the degree of smoothness penalty. The smoothing turned out to be insufficient in our numerical experiment, where the estimated $L(i,j)$ was still significantly affected by $F(i,j)$. Simply increasing smoothness penalty $\lambda$ had not solved this issue. To be more specific, we looked at the optimal solution of \eqref{eq:smooth} for an example electron microscope image, while varying $\lambda$ from 1 to 1000000. Figure \ref{fig:smooth_spline} shows one row of ground truth $L(i,j)$ and the same rows of the solution of \eqref{eq:smooth} with different $\lambda$'s. As shown in Fig. \ref{fig:smooth_spline}-(a), the row contains a mild slope in the background (red line), while foreground objects make significant intensity ditches on the slope. Applying the smoothing spline \eqref{eq:smooth} with a small $\lambda$ is led to an overfit to deep foreground ditches as shown in Fig. \ref{fig:smooth_spline}-(b) and -(c). Increasing $\lambda$ incurs a huge bias from the true background slope; see Fig. \ref{fig:smooth_spline}-(d). To reduce the effect of the foreground-caused-intensity-ditches on a background estimate, we borrow the concept of a robust regression in statistics \cite{rousseeuw2005robust}. In Section \ref{sec:form}, we describe how we formulate the background estimation problem, and the solution approach is described in Section \ref{sec:boosting}.

\subsubsection{Formulation} \label{sec:form}
The basic idea of our formulation is as follows. We regards foreground-caused intensity ditches as outliers deviating from a smooth background intensity surface. Estimating a smooth background intensity surface is formated as a robust regression problem that fits a regression model to an input image with minimal effect of the outliers. In the statistical literature \cite{rousseeuw2005robust}, the square loss criterion $(Y(i,j) - L(i,j))^2$ for fitting $L(i,j)$ is known to be sensitive to outliers. This is because the square loss is quickly increasing as absolute error $|Y(i,j) - L(i,j)|$ increases, so the regression fitting procedure that minimizes the square loss is prone to overfitting to outliers to avoid a huge surge of the square loss. In the robust statistics literature \cite{rousseeuw2005robust}, more robust loss functions were proposed in the form of the weighted square loss, 
\begin{equation*}
\rho_H(Y(i,j), L(i,j)) =  W(i,j)(Y(i,j) - L(i,j))^2,
\end{equation*}
where the weighting factor $W(i,j)$ is defined to lower weights on the $(i,j)$'s that outliers locate; smaller weights on outlier regions makes the outcome of the regression less affected by outlying features. A popular choice for the weight factor is the Huber loss weight \cite{rousseeuw2005robust},
\begin{equation*}
W(i,j) = \begin{cases}
1 & \mbox{ for } |Y(i,j) - L(i,j)| \le \delta \\
\frac{\delta}{|Y(i,j) - L(i,j)|} & \mbox{ otherwise,}
\end{cases}
\end{equation*}
where $\delta = 1.346$ is a popular choice. The Huber loss places lower weights for higher absolute difference $|Y(i,j) - L(i,j)|$, so the effect of extreme outliers on the regression estimate can be mitigated. We use the Huber loss to formulate a robust regression for a background surface intensity,
\begin{equation} \label{eq:rsmooth}
\min \sum_{i=1}^m \sum_{j=1}^n \rho_H(Y(i,j), L(i,j)) + \lambda \sum_{i=2}^{m-1} \sum_{j=2}^{n-1} |\nabla^2 L(i,j)|^2.
\end{equation} 
In the next section, we describe our modeling choice of $L(i,j)$ and our solution approach for problem \eqref{eq:rsmooth}.

\subsubsection{Solution Approach: Boosting Regression} \label{sec:boosting}
We model the background intensity surface $L(i,j)$ as an additive model of products of two one-dimensional functions,
\begin{equation} \label{eq:adm}
L(i,j) = U_1(i) \times V_1(j) + U_2(i) \times V_2(j) + \ldots + U_K(i) \times V_K(j),
\end{equation}
where $U_k(i)$ is a one-dimensional function of row index $i$, and $V_k(j)$ is a one-dimensional function of column index $j$. This model has much less degrees of freedom than the full matrix of $L(i,j)$; The degrees of freedom of the additive model is at most $K \times(M + N)$ when a degree of freedom is placed every $i$ for $U_k$ and is placed every $j$ for $V_k$, while the size of $L(i,j)$ is $M \times N$. Depending on the choice of $K$, it can model a simple background or a very complex background. For the time being, we assume $K$ is fixed, and we will later explain how $K$ can be chosen. \\

We solve problem \eqref{eq:rsmooth} with $L(i,j)$ in the additive form of \eqref{eq:adm}. An additive form of a regression function can be naturally fit by the boosting regression \cite[Chapter 10]{friedman2001elements}. Based on the boosting regression procedure, we devise Algorithm \ref{alg:boosting}. It starts with $L(i,j)=0$  and sequentially expands $L(i,j)$ with new functions through $K$ stages. Let $L^{(k-1)}(i,j)$ denote the result of the boosting regression after the $(k-1)$th stage. At the $k$th stage, a new product function $U_k(i) \times V_k(j)$ is added to $L^{(k-1)}(i,j)$ such that $L^{(k-1)}(i,j)$ plus the added term minimizes the objective function of \eqref{eq:rsmooth} without changing the other terms added in the previous stages, 
\begin{equation*} 
\begin{split}
(U_k, V_k) = \arg\min_{U, V} \quad  & \sum_{i=1}^m \sum_{j=1}^n \rho_H(Y(i,j), L^{(k-1)}(i,j) + U(i) \times V(j)) \\
                   & + \lambda \sum_{i=2}^{m-1} \sum_{j=2}^{n-1} |\nabla^2 (U \times V)(i,j)|^2,
\end{split}
\end{equation*}
where $\nabla^2 (U \times V)(i,j)$ is the $2$nd order derivative of $U \times V$ at $(i,j)$. Define $R^{(k)}(i,j) = Y(i,j) - L^{(k-1)}(i,j)$ as the residual of fit after the $(k-1)$th stage. The $k$th stage is basically equivalent to fitting $U_k(i) \times V_k(j)$ to the residual,
\begin{equation} \label{eq:stage}
\begin{split}
(U_k, V_k) = \arg\min_{U, V} \sum_{i=1}^m \sum_{j=1}^n & \rho_H(R^{(k)}(i,j), U \times V ) \\
                                      &+ \lambda \sum_{i=2}^{m-1} \sum_{j=2}^{n-1} |\nabla^2 (U \times V)(i,j)|^2.
\end{split}
\end{equation} 
Once $U_k$ and $V_k$ are determined, the update $L^{(k)}(i,j) = L^{(k-1)}(i,j) + U_k(i) \times V_k(j)$ is performed as the last step of the $k$th stage. The number of the stages performed determines the number of product function terms in the additive model. We use the following stopping criterion to determine when we stop the sequential addition, 
\begin{equation*}
||U_k(i)||_2^2 \cdot  ||V_k(j)||_2^2 \le \epsilon,
\end{equation*}
where $||U_k(i)||_2^2$ is the L2-norm of a function that is approximated by its discrete version $\sum_{i=1}^m U_k(i)^2$, and similarly $||V_k(j)||_2^2 =\sum_{j=1}^n V_k(i)^2$. The stopping criterion practically implies that adding additional terms to the additive model is unnecessary when the last term added is ignorable. The overall algorithm is described in Algorithm \ref{alg:boosting}. \\
\begin{algorithm*}
\caption{Boosting Regression} \label{alg:boosting}
\begin{algorithmic}[1]
\REQUIRE input image $Y(i,j)$
\ENSURE background intensity surface $L(i,j)$
\STATE \textbf{Initialization:} residual $R^{(0)}(i,j) = Y(i,j)$ and initial background intensity surface $L(i,j) = 0$
\FOR{$k=1$ to $K$}
\STATE Fit $U_k(i) \times V_k(j)$ to $R^{(k)}(i,j)$, based on optimization \eqref{eq:rsmooth}.\\
\indent\hspace{0.5cm} [\textit{This optimization is solved by Algorithm} \ref{alg:opt}.] 
\STATE $R^{(k+1)}(i,j) = R^{(k)}(i,j)-U_k(i)\times V_k(j)$
\STATE $L(i,j) = L(i,j) + U_k(i) \times V_k(j)$
\STATE If $\sum_i U_k(i)^2 \sum_j V_k(j)^2 < \epsilon$, stop and otherwise continue.
\ENDFOR
\end{algorithmic}
\end{algorithm*}

The remainder of this section is focused on detailing how to solve each stage formulated as problem \eqref{eq:stage}, i.e., line 3 of Algorithm \ref{alg:boosting}. We first model $U(i)$ as a smooth function that interpolates discrete points $\{(i, u_i); i=1,2,\ldots,m\}$, where the term `smooth function'  implies that the second order derivative of a function has a bounded magnitude. Therefore, $U(i)  = u_i$, and the first and second order derivatives of $U(i)$ are approximated by the central difference approximation of the first and second derivatives of $U(i)$, 
\begin{equation*}
\begin{split}
& U'(i) = (U(i+1) - U(i-1))/2 = (u_{i+1} - u_{i-1})/2, \\
& U''(i) = U(i-1) - 2 U(i) + U(i+1) = u_{i-1} - 2 u_i + u_{i+1}.
\end{split}
\end{equation*}
Similarly, we model $V(j)$ as a smooth function that interpolates discrete points $\{(j, v_j); j=1,2,\ldots,n\}$, so
\begin{equation*}
\begin{split}
& V(j)  = v_j, \\
& V'(j) = (V(j+1) - V(j-1))/2 = (v_{j+1} - v_{j-1})/2, \\
& V''(j) = V(j-1) - 2 V(j) + V(j+1) = v_{j-1} - 2 v_j + v_{j+1}.
\end{split}
\end{equation*}
With the modeling of $U(i)$ and $V(j)$, we can restate the robust loss in problem \eqref{eq:stage} as follows:
\begin{equation} \label{eq:rloss}
\begin{split}
\rho_H(R^{(k)}(i,j),& U(i)\times V(j)) \\ 
              &= W(i,j) (R^{(k)}(i,j) - u_iv_j)^2,
\end{split}
\end{equation}
and the smoothness term in the objective function of \eqref{eq:stage} is the Frobenius norm of the Hessian matrix of $(U \times V)(i,j)$, 
\begin{equation} \label{eq:spen}
\begin{split}
|\nabla^2 (U \times V)(i,j)|^2 = & U''(i)^2 V(j)^2 + U(i)^2 V''(j)^2 \\
                                  &+ 2U'(i)^2 V'(j)^2 \\
                                = & (u_{i-1} - 2 u_i + u_{i+1})^2 v_j^2 \\
                                  & + u_i^2 (v_{j-1} - 2 v_j + v_{j+1})^2 \\
                                  & + 2\left(\frac{u_{i+1} - u_{i-1}}{2} \right)^2 \left(\frac{v_{j+1} - v_{j-1}}{2} \right)^2.
\end{split}
\end{equation}
Combining \eqref{eq:rloss} and \eqref{eq:spen}, we can rewrite the objective function of problem \eqref{eq:stage} as
\begin{equation} \label{eq:obj} 
\begin{split}
\sum_{i=1}^{m}\sum_{j=1}^{n} & W(i,j) (R^{(k)}(i,j) - u_iv_j)^2 \\
                                  & + \sum_{i=2}^{m-1} (u_{i-1} - 2 u_i + u_{i+1})^2 \sum_{j=2}^n v_j^2 \\
                                  & + \sum_{i=2}^{m-1} u_i^2 \sum_{j=2}^{n-1} (v_{j-1} - 2 v_j + v_{j+1})^2 \\
                                  & + 2\sum_{i=2}^{m-1} \left(\frac{u_{i+1} - u_{i-1}}{2} \right)^2 \sum_{j=2}^{n-1}\left(\frac{v_{j+1} - v_{j-1}}{2} \right)^2.
\end{split}
\end{equation}
Let us simplify the expression using some vector and matrix notations. Let $\V{u} = (u_1, u_2, \ldots, u_m)^T$ and  $\V{v} = (v_1, v_2, \ldots, v_n)^T$. Let $\M{R}^{(k)}$ denote the $m \times n$ matrix with $R^{(k)}(i,j)$ as its $(i,j)$th element, and let $\M{W}^{1/2}$ denote the $m \times n$ matrix with $\sqrt{W(i,j)}$ as its $(i,j)$th element. We also introduce a $m \times m$ matrix $\M{\Omega}_m$ to represent the quadratic term $\sum_{i=2}^{m-1} (u_{i-1} - 2 u_i + u_{i+1})^2$ as $\V{u}^T \M{\Omega}_m \V{u}$, and introduce another $m \times m$ matrix $\M{\Gamma}_m$ to represent the quadratic term $\sum_{i=2}^{m-1} ((u_{i+1} - u_{i-1})/2)^2$ as $\V{u}^T \M{\Gamma}_m \V{u}$. Similarly, we introduce two $n \times n$ matrices, $\M{\Omega}_n$ and $\M{\Gamma}_n$, for representing the quadratic terms $\sum_{j=2}^{n-1} (v_{j-1} - 2 v_j + v_{j+1})^2$ and $\sum_{j=2}^{n-1}((v_{j+1} - v_{j-1})/2)^2$ respectively. The vectorial form of \eqref{eq:obj} is
\begin{equation*}
\begin{split}
f(\V{u},\V{v}; \lambda) =  & ||\M{W}^{1/2} \circ (\M{R}^{(k)} - \V{u}\V{v}^T)||_F^2 \\
& +\lambda \V{u}^T \M{\Omega}_m \V{u} \cdot \V{v}^T \V{v} \\
& + \lambda \V{v}^T \M{\Omega}_n \V{v} \cdot  \V{u}^T \V{u} \\
& + 2 \lambda \V{u}^T \M{\Gamma}_m \V{u} \cdot \V{v}^T \M{\Gamma}_n \V{v},
\end{split}
\end{equation*}
where $\circ$ is the Hadamard product operator and $||\cdot||_F$ is the Frobenius norm. \\

Problem \eqref{eq:stage} that minimizes $f(\V{u},\V{v}; \lambda)$ for $\V{u}$ and $\V{v}$ can be solved by the coordinate descent algorithm, which iterates two steps: (a) optimizing $\V{u}$ while fixing $\V{v}$ and (b) optimizing $\V{v}$ while fixing $\V{u}$. We first derive the optimization procedure for $\V{u}$ with fixed $\V{v}$. Please note that the Frobenius norm of an arbitrary matrix $\M{X}$ is 
\begin{equation*}
\begin{split}
||\M{X}||_F^2 & = \vec(\M{X})^T \vec(\M{X}),
\end{split}                                              
\end{equation*}
where $\vec(\M{X})$ is the vectorization of the matrix $\M{X}$. The vectorization of $\M{W}^{1/2} \circ (\M{R}^{(k)} - \V{u}\V{v}^T)$ is  
\begin{equation*}
\begin{split}
\vec(\M{W}^{1/2} & \circ (\M{R}^{(k)} - \V{u}\V{v}^T)) \\
   & = \vec(\M{W}^{1/2}) \circ (\vec(\M{R}^{(k)}) - (\V{v} \otimes \V{I}_m) \V{u}) \\
   & = diag(\vec(\M{W}^{1/2}))(\vec(\M{R}^{(k)}) - (\V{v} \otimes \V{I}_m) \V{u}), 
\end{split}                                              
\end{equation*}
where $\otimes$ is the Kronecker product, $\M{I}_m$ is an identity matrix of size $m$, and $diag(\V{v})$ is a diagonal matrix with the elements of $\V{v}$ as its diagonal elements. Let $\mathcal{W}^{1/2} = diag(\vec(\M{W}^{1/2}))$, $\V{r}_k = \vec(\M{R}^{(k)})$ and $\mathcal{V} = (\V{v} \otimes \V{I}_m)$. The previous expression is restated as 
\begin{equation*}
\begin{split}
\vec(\M{W}^{1/2} & \circ (\M{R}^{(k)} - \V{u}\V{v}^T)) \\
   & = \mathcal{W}^{1/2}(\V{r}_k - \mathcal{V}\V{u}),
\end{split}                                              
\end{equation*}
and the following Frobenius norm is  
\begin{equation*}
\begin{split}
||\M{W}^{1/2} \circ (\M{R}^{(k)} - \V{u}\V{v}^T)||_F^2 =(\V{r}_k - \mathcal{V}\V{u})^T \mathcal{W} (\V{r}_k - \mathcal{V}\V{u}),      
\end{split}                                              
\end{equation*}
where $\mathcal{W} = \mathcal{W}^{1/2}\mathcal{W}^{1/2}$. The partial derivative of $f(\V{u},\V{v}; \lambda)$ with respect to $\V{u}$ is 
\begin{equation*}
\begin{split}
\frac{\partial f(\V{u}, \V{v}; \lambda)}{\partial \V{u}} = & -2\mathcal{V}^T \mathcal{W} (\V{r}_k - \mathcal{V}\V{u}) \\
                                                  & + 2\lambda  \M{\Omega}_{u|v} \V{u},
\end{split}
\end{equation*}
where $\M{\Omega}_{u|v} = (\V{v}^T \V{v} \M{\Omega}_m + \V{v}^T \M{\Omega}_n \V{v} \M{I}_m + 2 \V{v}^T \M{\Gamma}_n \V{v} \M{\Gamma}_m )$. Solving $\frac{\partial f(\V{u}, \V{v})}{\partial \V{u}} = 0$ for $\V{u}$ gives the optimal solution for $\V{u}$ when $\V{v}$ is fixed,
\begin{equation*}
\begin{split}
\V{u} = (\mathcal{V}^T \mathcal{W} \mathcal{V} + \lambda \M{\Omega_{u|v}})^{-1} \mathcal{V}^T \mathcal{W}\V{r}_k.
\end{split}                                              
\end{equation*}
\\
Similarly, the update procedure for $\V{v}$ can be derived. Let $\mathcal{U} = (\V{I}_n \otimes \V{u})$. Using the properties of the vectorization, $\vec(\M{W}^{1/2} \circ (\M{R}^{(k)} - \V{u}\V{v}^T))$ can be restated for $\V{v}$ as
\begin{equation*}
\begin{split}
\vec(\M{W}^{1/2} \circ (\M{R}^{(k)} - \V{u}\V{v}^T)) = \mathcal{W}^{1/2} (\V{r}_k - \mathcal{U}\V{v}),
\end{split}                                              
\end{equation*}
and 
\begin{equation*}
\begin{split}
||\M{W}^{1/2} \circ (\M{R}^{(k)} - \V{u}\V{v}^T)||_F^2 = (\V{r}_k - \mathcal{U}\V{v})^T \mathcal{W} (\V{r}_k - \mathcal{U}\V{v}).
\end{split}                                              
\end{equation*}
The partial derivatives of $f(\V{u},\V{v}; \lambda)$ with respect to $\V{v}$ are
\begin{equation*}
\begin{split}
\frac{\partial f(\V{u}, \V{v}; \lambda)}{\partial \V{v}} = & -2\mathcal{U}^T \mathcal{W} (\V{r}_k - \mathcal{U}\V{v}) \\
                                                  & + 2 \lambda \M{\Omega}_{v|u} \V{v},
\end{split}
\end{equation*}
where $\M{\Omega}_{v|u} = (\V{u}^T \M{\Omega}_m \V{u} \M{I}_n + \V{u}^T \V{u} \M{\Omega}_n + 2 \V{u}^T \M{\Gamma}_m \V{u} \M{\Gamma}_n)$. Solving $\frac{\partial f(\V{u}, \V{v})}{\partial \V{v}} = 0$ gives the optimal solution for $\V{v}$ when $\V{u}$ is fixed, 
\begin{equation*}
\begin{split}
\V{v} = (\mathcal{U}^T \mathcal{W} \mathcal{U} + \lambda\M{\Omega_{v|u}})^{-1} \mathcal{U}^T \mathcal{W}\V{r}_k.
\end{split}                                              
\end{equation*}
\\
Once both of $\V{u}$ and $\V{v}$ is updated, the weighting factor $\mathcal{W}$ is updated using 
\begin{equation} \label{eq:W}
W(i,j) = \begin{cases}
1 & \mbox{ for } |R^{(k)}(i,j) - u_i v_j| \le \delta \\
\frac{\delta}{|R^{(k)}(i,j) - u_i v_j|} & \mbox{ otherwise.}
\end{cases}
\end{equation}
Updating $\V{u}$, $\V{v}$ and $W(i,j)$'s are repeated until convergence. The details of the algorithm is summarized in Algorithm \ref{alg:opt}.\\

\begin{algorithm*}
\caption{Iterative Optimization} \label{alg:opt}
\begin{algorithmic}[1]
\REQUIRE residual $\M{R}^{(k)}$, $\lambda$, $\epsilon$.
\ENSURE $\V{u}$ and $\V{v}$
\STATE \textbf{Initialization:} $\V{r}_k = vec(\M{R}^{(k)})$. Set $\V{u}_{old} = \V{0}$ and $\V{v}_{old} = \V{0}$. \\Set $\V{u}$ and $\V{v}$ with the first left and right singular vectors of $\M{R}^{(k)}$.  
\WHILE{$||\V{u}_{old}\V{v}_{old}^T - \V{u}\V{v}^T||_F^2 > \epsilon$}
\STATE Update $\mathcal{W}$ using the formula \eqref{eq:W}.
\STATE Update $\M{\Omega}_{u|v} = \V{v}^T \V{v} \M{\Omega}_m + \V{v}^T \M{\Omega}_n \V{v} \M{I}_m + 2 \V{v}^T \M{\Gamma}_n\V{v} \M{\Gamma}_m$ and $\mathcal{V} = \V{v} \otimes \M{I}_m$. 
\STATE Compute $\V{u}_{old} = \V{u}$ and $\V{u} = (\mathcal{V}^T \mathcal{W} \mathcal{V} + \lambda \M{\Omega_{u|v}})^{-1} \mathcal{V}^T \mathcal{W}\V{r}_k$.
\STATE Update $\M{\Omega}_{v|u} = \V{u}^T \M{\Omega}_m \V{u} \M{I}_n + \V{u}^T \V{u} \M{\Omega}_n + 2 \V{u}^T \M{\Gamma}_m \V{u} \M{\Gamma}_n$ and $\mathcal{U} = \M{I}_n \otimes \V{u}$.
\STATE Compute $\V{v}_{old} = \V{v}$ and $\V{v} = (\mathcal{U}^T \mathcal{W} \mathcal{U} + \lambda \M{\Omega_{v|u}})^{-1} \mathcal{U}^T \mathcal{W}\V{r}_k.$
\ENDWHILE
\end{algorithmic}
\end{algorithm*}
The proposed algorithm has a tuning parameter $\lambda$ that determines the smoothness of the background estimation. We run Algorithm \ref{alg:opt} with different values of $\lambda$. Let $\V{u}_{\lambda}$ and $\V{v}_{\lambda}$ denote the outputs of the algorithm with a choice of $\lambda$. We choose $\lambda$ based on the criterion,
\begin{equation*}
\min_{\lambda} f(\V{u}_{\lambda}, \V{v}_{\lambda}; \lambda).
\end{equation*}
The values of $\lambda$ that we considered are $\{10^{-4}, 10^{-2}, 10^{0}, 10^{2}, 10^4\}$.\\

\begin{figure}
\centering
\includegraphics[width=\textwidth]{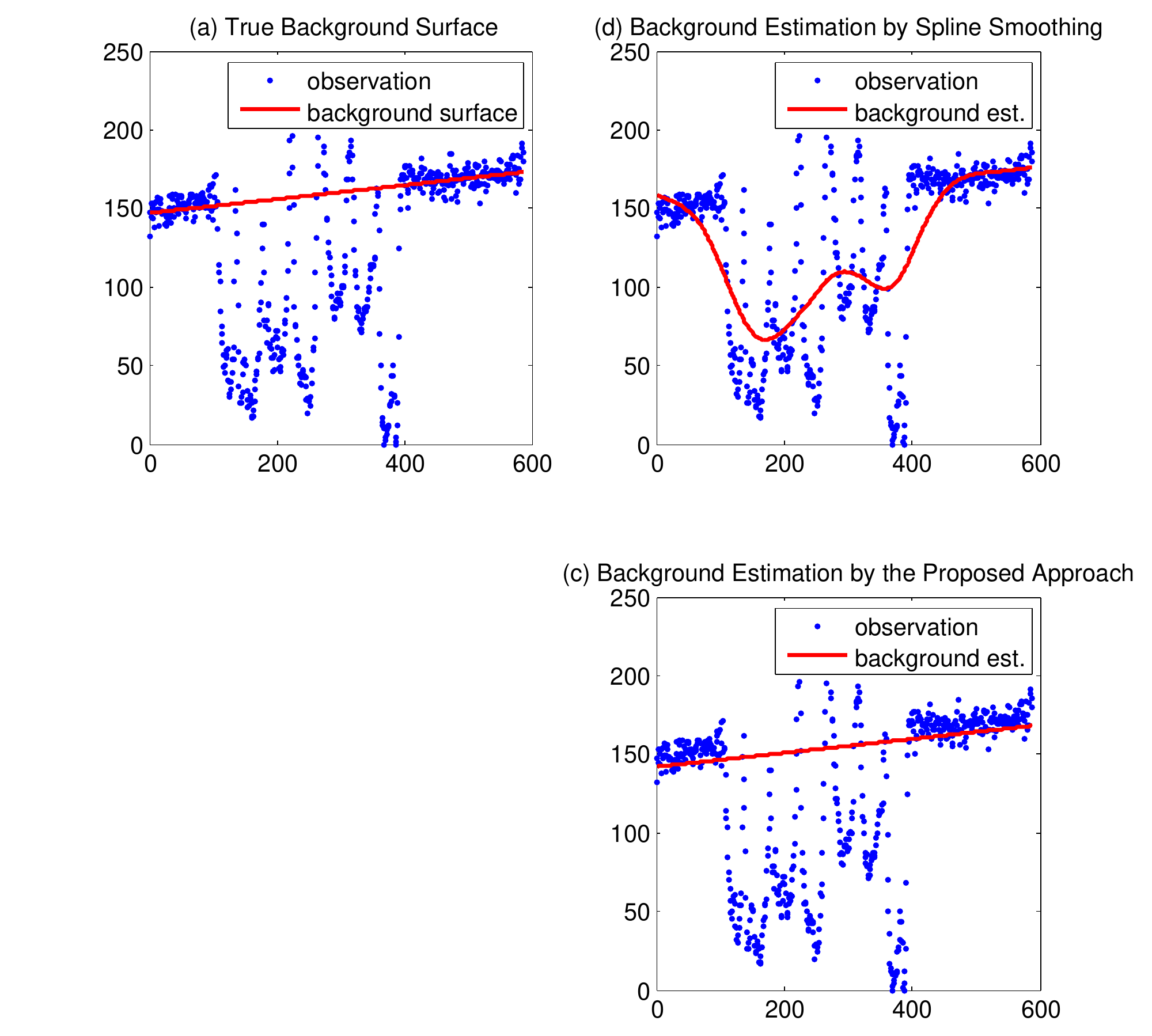}
\caption{Illustrative Comparison. Smoothing Spline Regression vs Proposed Approach} \label{fig:smooth_spline2}
\end{figure}

The proposed algorithm was applied for the example data used in Figure \ref{fig:smooth_spline}. It worked very well as illustrated in Figure \ref{fig:smooth_spline2}. More examples can be found in Section \ref{sec:bin_outcome}.

\subsection{Choosing $\tau$} \label{sec:fgr}
This section describes how to choose a global threshold $\tau$ that binarizes the background subtracted image $Y(i,j) - L(i,j)$ as follows: 
\begin{equation*}
B(i,j) = \begin{cases} 
1 & \mbox{ for } Y(i,j) - L(i,j) \le \tau\\
0 & \mbox{ for } Y(i,j) - L(i,j) > \tau.
\end{cases}
\end{equation*}
Once the background subtraction is subtracted, one may consider to apply the popular global threshold selector on $Y(i,j) - L(i,j)$ such as Otsu \cite{otsu1979threshold}. The Otsu threshold selector is based on the histogram of $Y(i,j) - L(i,j)$, and it works best when the histogram is bimodal. However, for most of our example images, the histograms looked unimodal, which is mainly because the number of foreground pixels is significantly dominated by the number of background pixels. We formulate the threshold selection problem as a model selection problem of a regression parameter and use a model selection criterion to select a threshold. \\

Let $\tilde{Y}(i,j) = Y(i,j) - L(i,j)$. From the literature \cite{friedman2001elements}, it is well known that the hard-thresholded image $\tilde{Y}(i,j) B(i,j)$ is the optimal solution for the $L0$-penalized regression problem,
\begin{equation*} \label{eq:gMDL}
\min_{F} \sum_{i=1}^{m}\sum_{j=1}^{n} (\tilde{Y}(i,j)-F(i,j))^2 + \tau^2 \sum_{i=1}^{m}\sum_{j=1}^{n} ||F(i,j)||_0.
\end{equation*}
Therefore, the threshold selection of $\tau$ can be recast as the model selection problem of L0 penalty parameter $\tau^2$. Many model selection criteria for a regularized regression were proposed, including the Akaike information criterion (AIC), Bayesian information criterion (BIC), $C_p$ statistics and model description length (MDL) \cite{friedman2001elements}. More recently, the generalized model description length (gMDL) was proposed \cite{hansen2003minimum}, and it takes a mixture form that can choose one in between two selection criteria, AIC and BIC, depending on which is best for data in hand. We use the gMDL criterion to select $\tau$. Let $RSS_{\tau} =  \sum_{i=1}^m \sum_{j=1}^n I(Y(i,j) > \tau) (\tilde{Y}(i,j))^2$, $FSS = \sum_{i=1}^m \sum_{j=1}^n (\tilde{Y}(i,j))^2$ and $p_{\tau} = \sum_{i=1}^m \sum_{j=1}^n I(Y(i,j) > \tau)$, where $I(\cdot)$ is the indicator function. Based on \cite[equation (16)]{hansen2003minimum}, the gMDL for problem \eqref{eq:gMDL} is defined as follows: if the coefficient of determination for $\tau^2$ is less than $mn/p_{\tau}$, $gMDL(\tau)$ is
\begin{equation*}
\frac{mn}{2} \log \frac{RSS_{\tau} }{mn - p_{\tau}} + \frac{p_{\tau} }{2} \log \frac{(mn-p_{\tau} )FSS}{p_{\tau} RSS_{\tau} } + \log (mn),
\end{equation*}
and $\frac{mn}{2} \log(\frac{FSS}{mn}) + \frac{1}{2}\log(mn)$ otherwise. We evaluated $gMDL(\tau)$ for each $\tau$ value in all unique values $\tilde{Y}(i,j)$'s, and choose one that achieves the lowest gMDL value.  

\section{Numerical Evaluation} \label{sec:app}
\begin{figure}
\centering
  \subfigure[image 1]{    
	\includegraphics[width=0.2\textwidth]{images/HW1.jpg}	
}
    \subfigure[image 2]{    
	\includegraphics[width=0.2\textwidth]{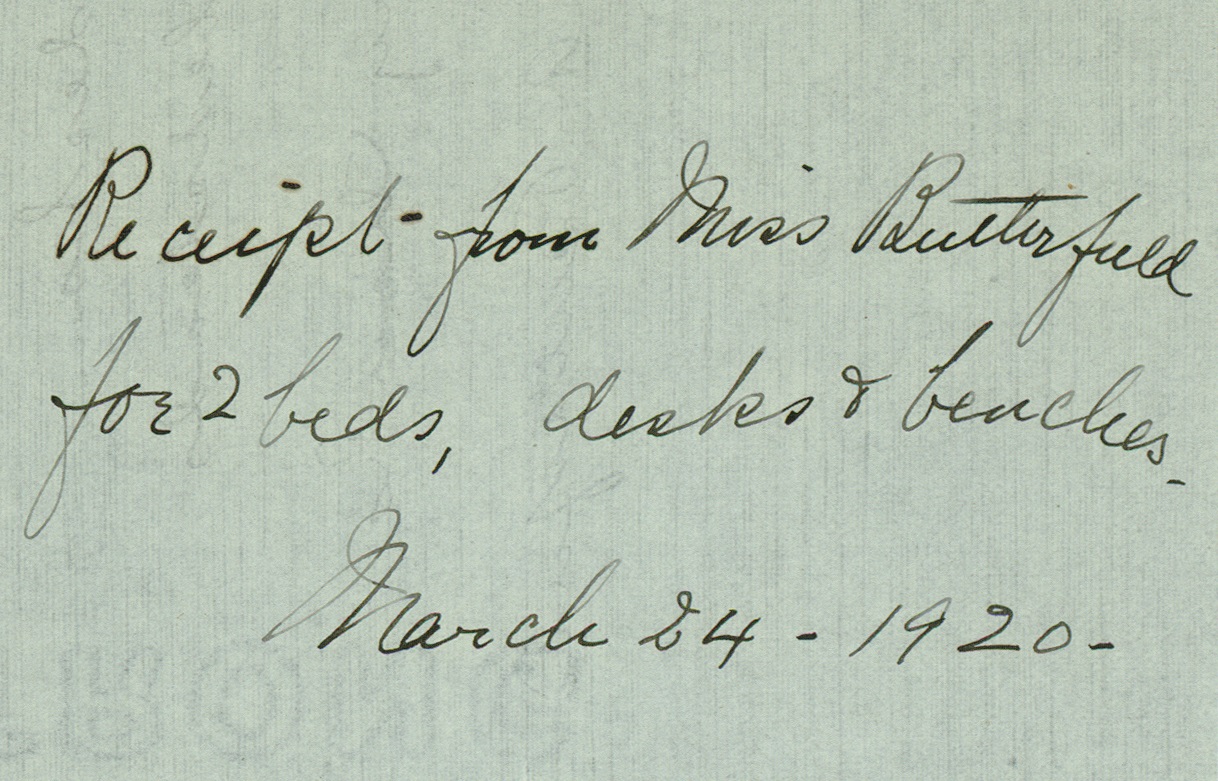}	
}   
  \subfigure[image 3]{    
	\includegraphics[width=0.2\textwidth]{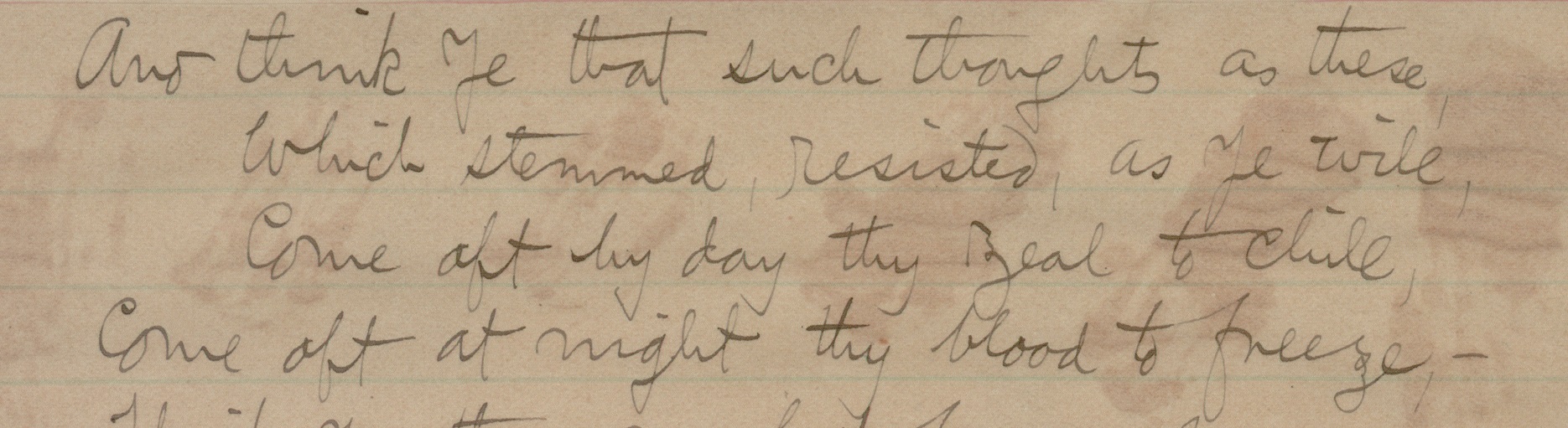}	
}
  \subfigure[image 4]{    
	\includegraphics[width=0.2\textwidth]{images/HW4.jpg}	
}   
  \subfigure[image 5]{    
	\includegraphics[width=0.2\textwidth]{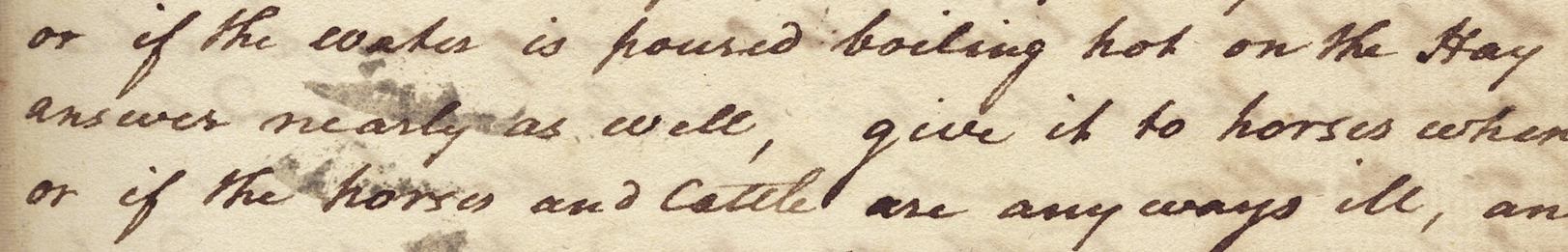}	
}
\subfigure[image 6]{
     	\includegraphics[width=0.2\textwidth]{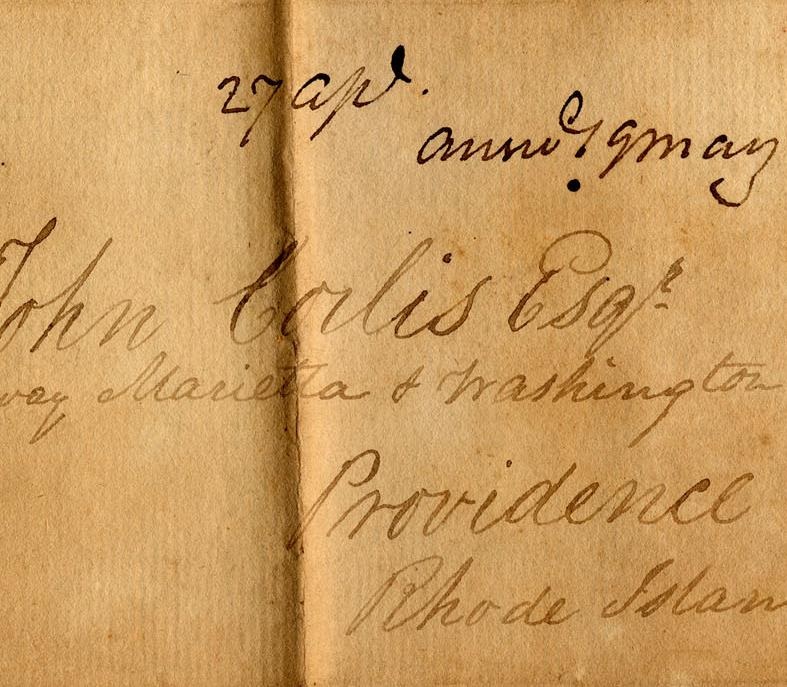}
}
\subfigure[image 7]{
     	\includegraphics[width=0.2\textwidth]{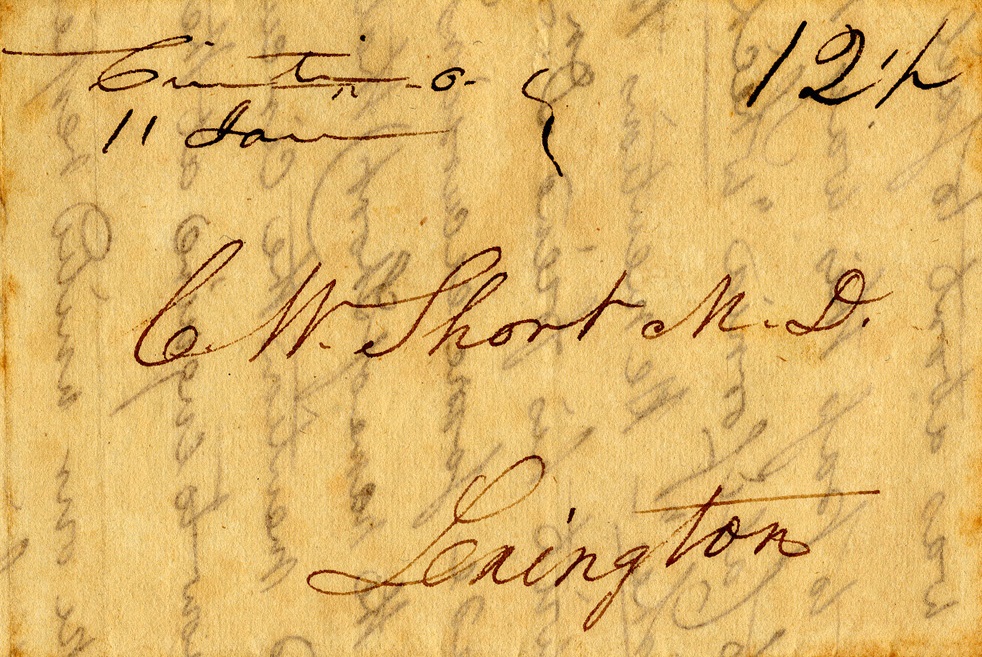}
}  
\subfigure[image 8]{
     	\includegraphics[width=0.2\textwidth]{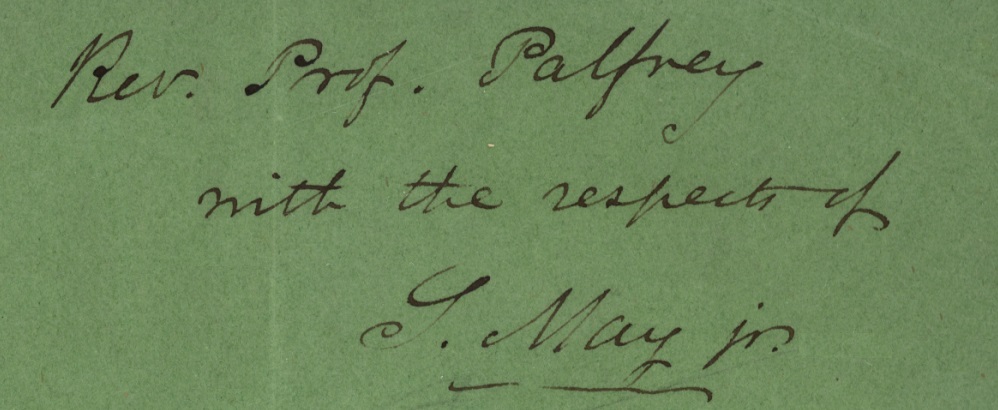}
}
\caption{\texttt{DIBCO11} Dataset}
\label{fig_dataset1}
\end{figure}

\begin{figure}
\centering
\centering
  \subfigure[image 1]{    
	\includegraphics[width=0.2\textwidth]{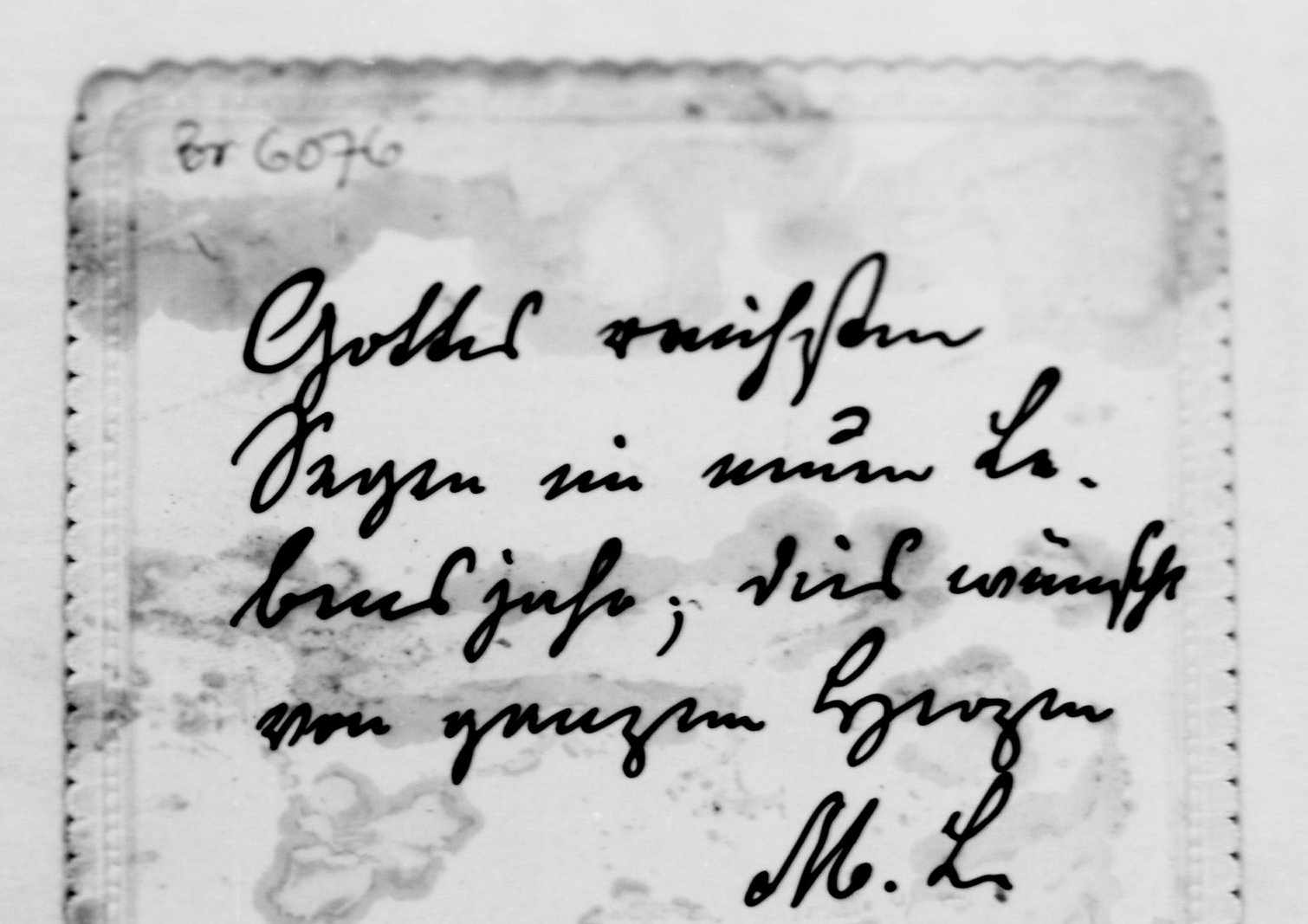}	
}
    \subfigure[image 2]{    
	\includegraphics[width=0.2\textwidth]{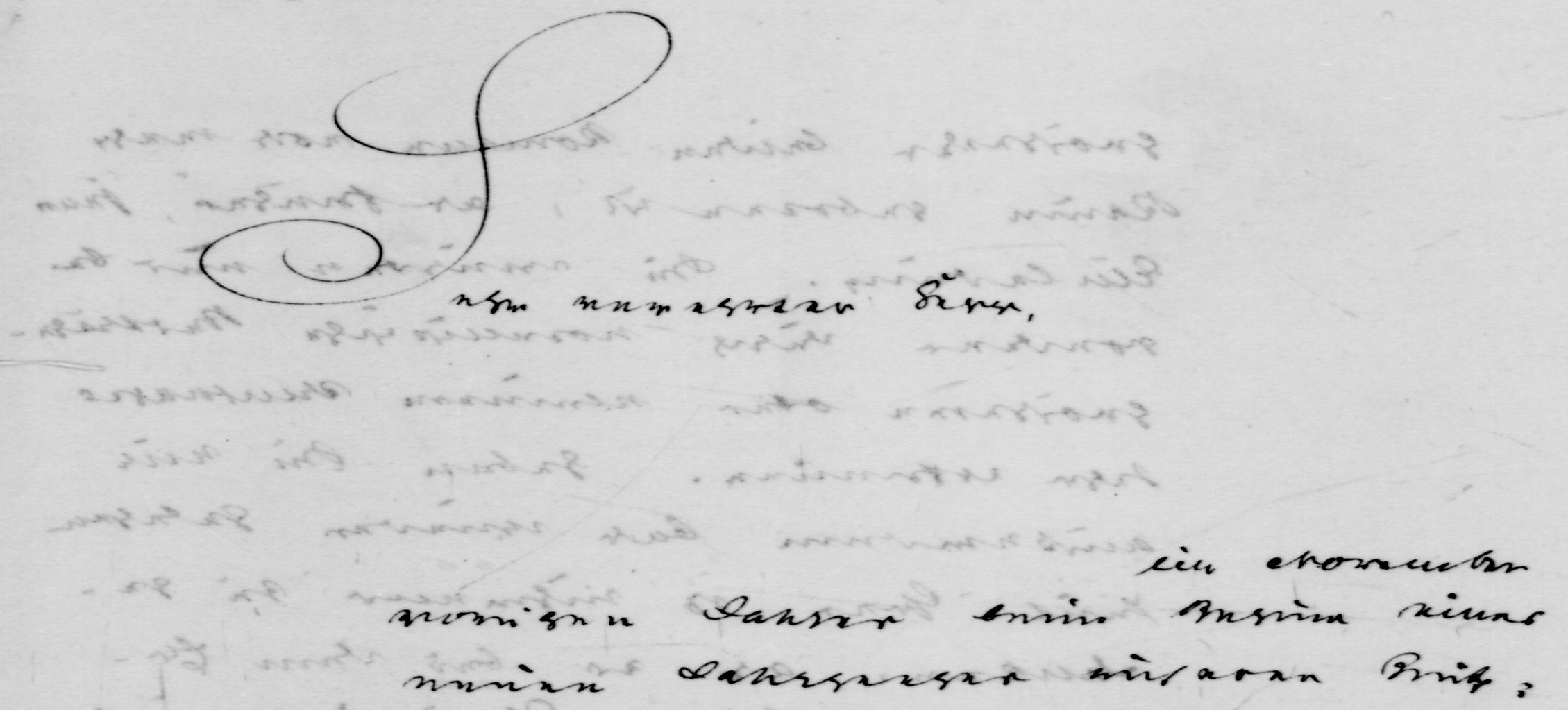}	
}   
  \subfigure[image 3]{    
	\includegraphics[width=0.2\textwidth]{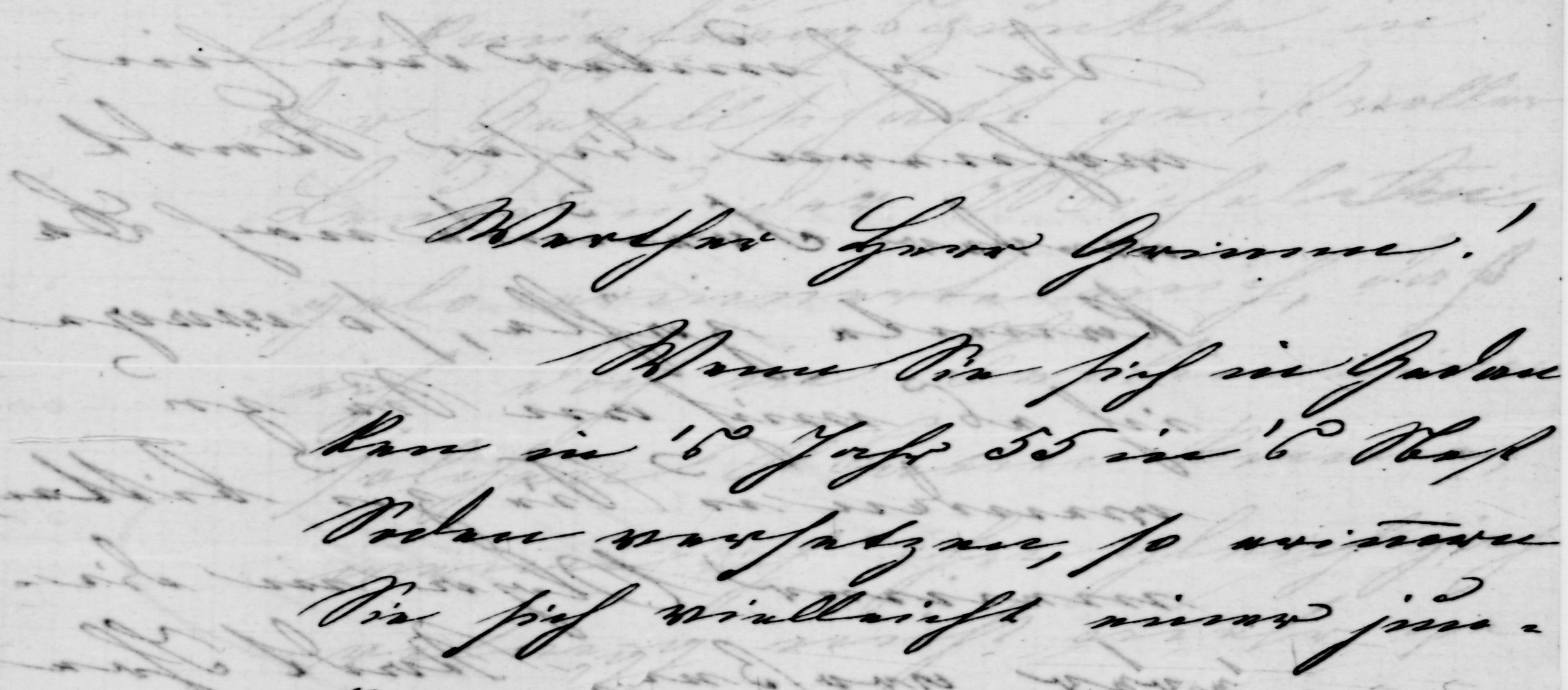}	
}
  \subfigure[image 4]{    
	\includegraphics[width=0.2\textwidth]{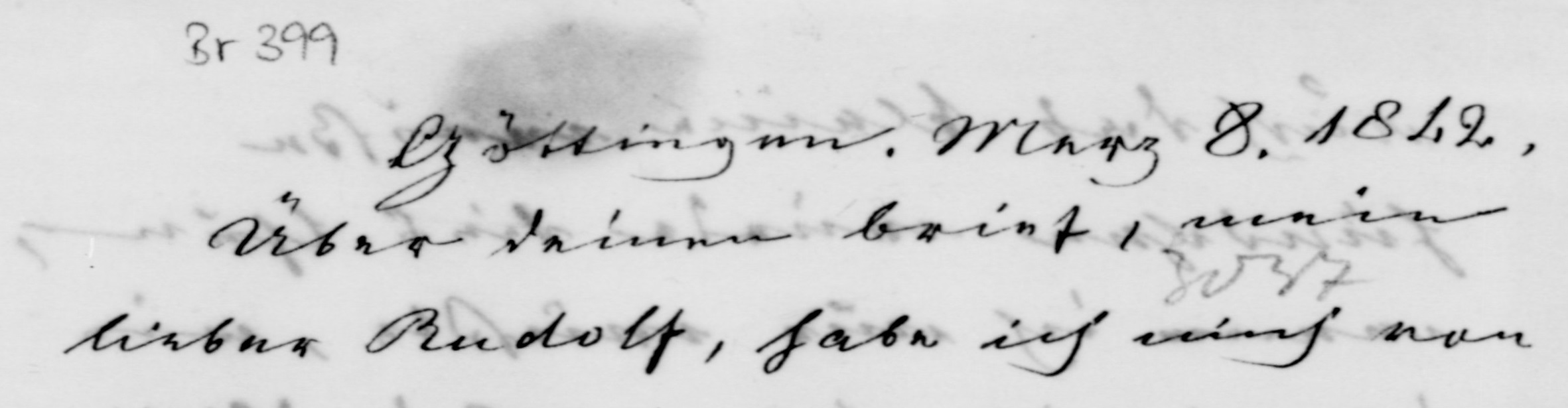}	
}   
  \subfigure[image 5]{    
	\includegraphics[width=0.2\textwidth]{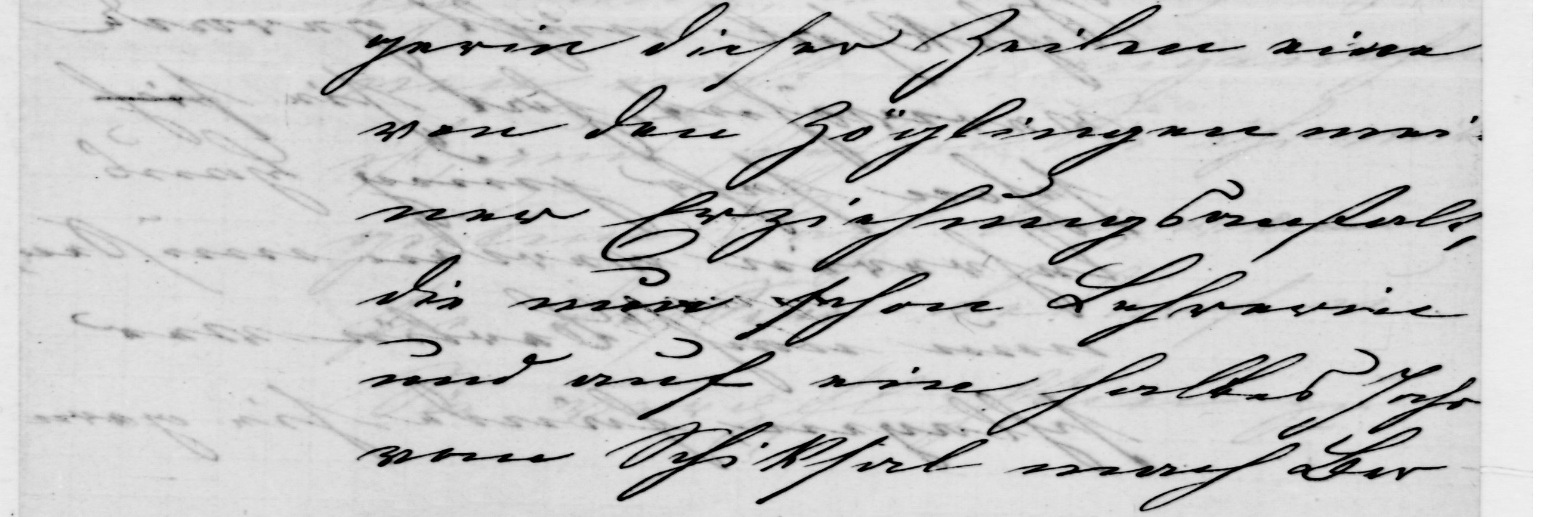}	
}   
\subfigure[image 6]{
     	\includegraphics[width=0.2\textwidth]{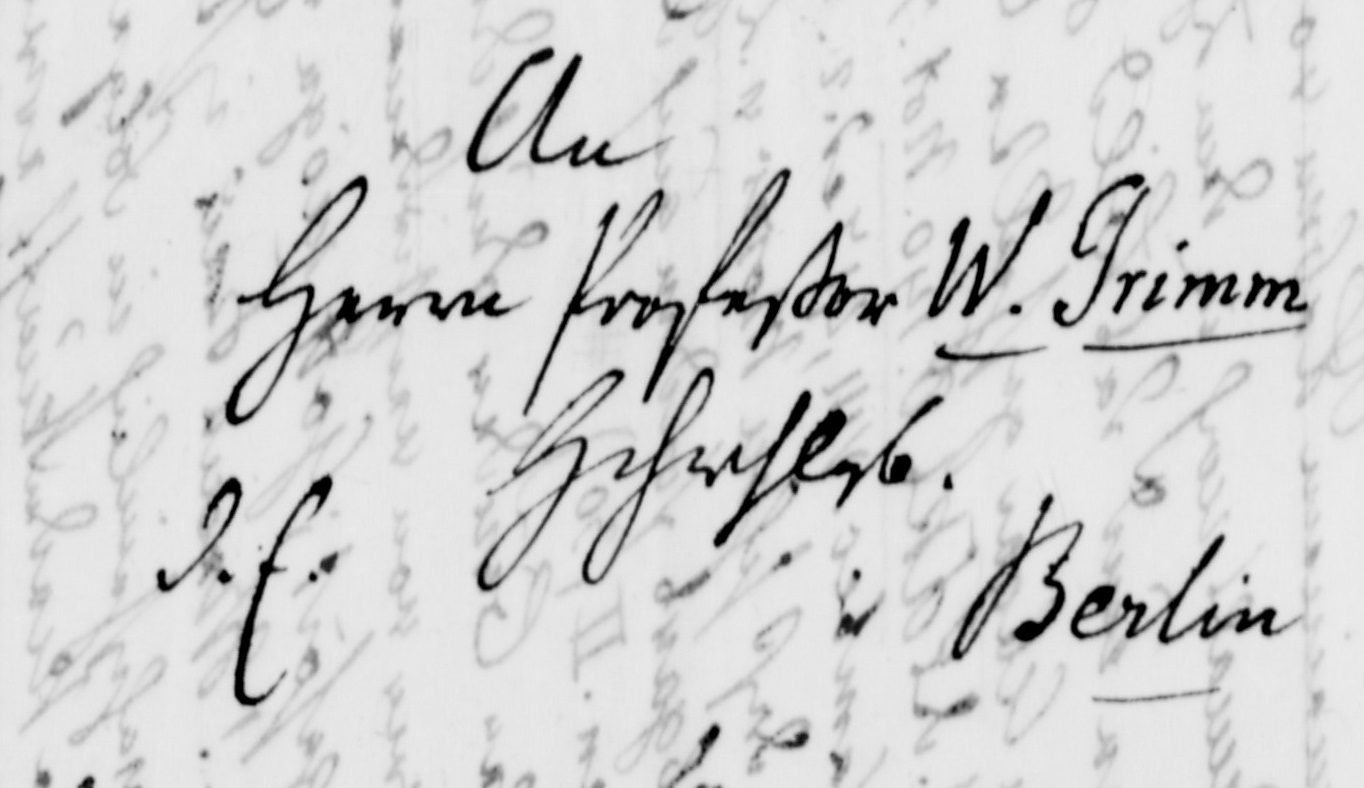}
} 
\subfigure[image 7]{
     	\includegraphics[width=0.2\textwidth]{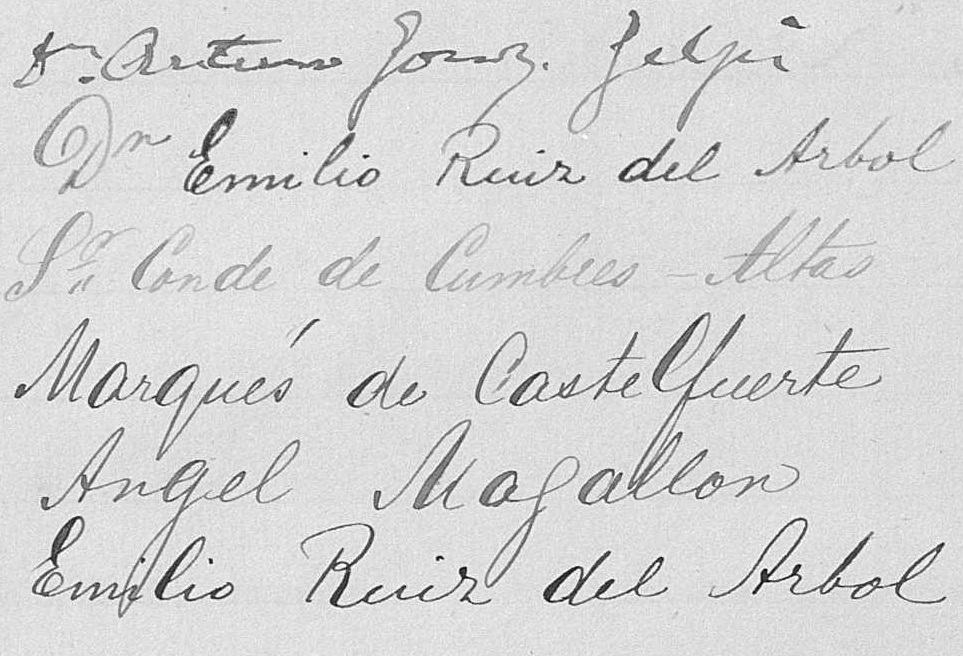}
}
\subfigure[image 8]{
     	\includegraphics[width=0.2\textwidth]{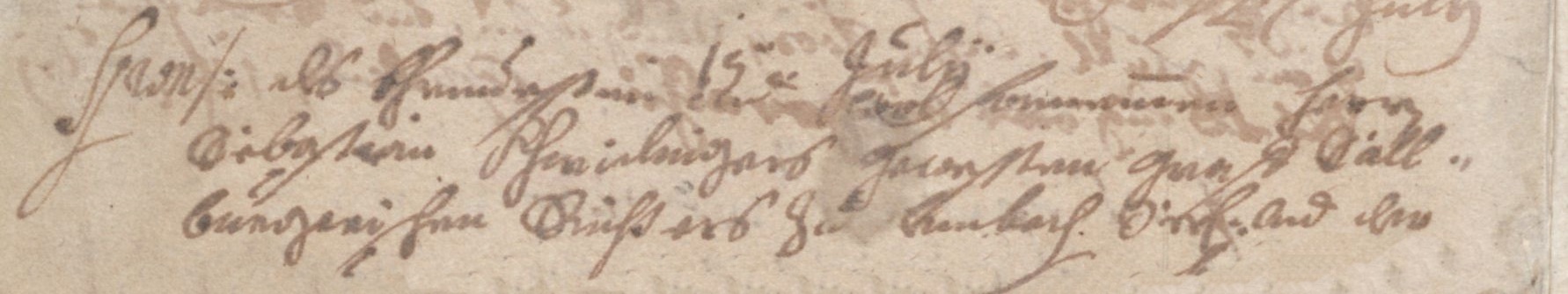}
}
\subfigure[image 9]{
     	\includegraphics[width=0.2\textwidth]{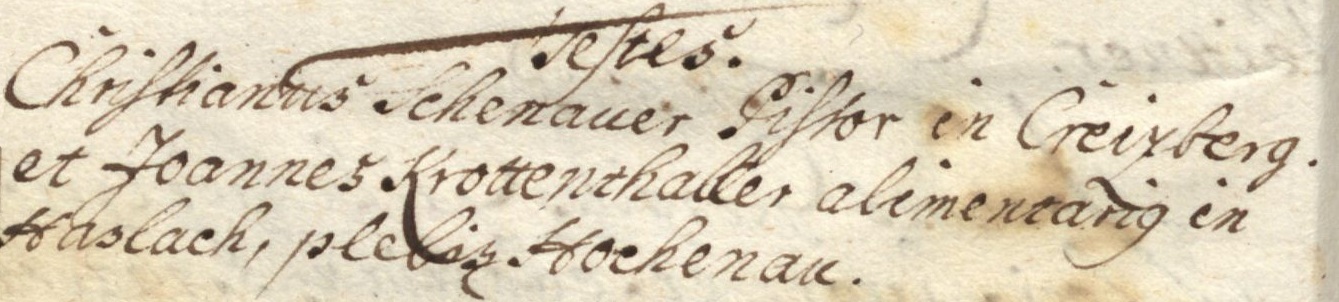}
}
\subfigure[image 10]{
     	\includegraphics[width=0.2\textwidth]{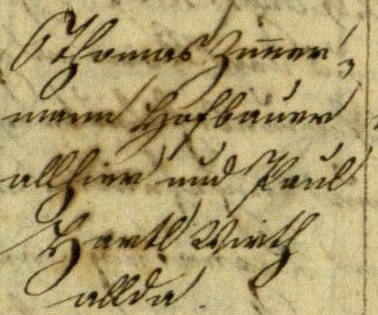}
}
\caption{\texttt{H-DIBCO16} Dataset}
\label{fig_dataset2}
\end{figure}

\begin{figure}
\centering
  \subfigure[image 1]{    
	\includegraphics[width=0.18\textwidth]{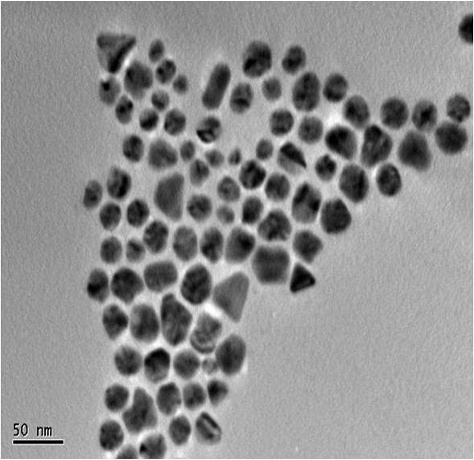}	
}
  \subfigure[image 2]{    
	\includegraphics[width=0.18\textwidth]{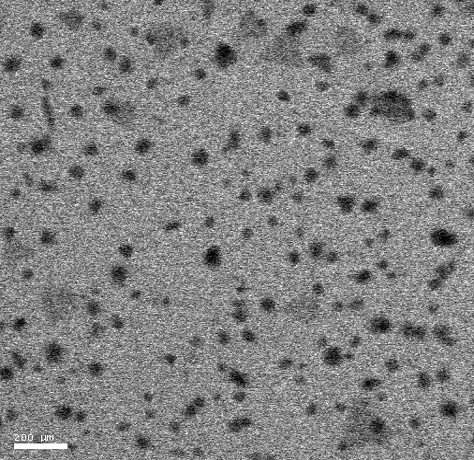}	
}
\subfigure[image 3]{
     	\includegraphics[width=0.18\textwidth]{images/Image151.jpg}
}
\subfigure[image 4]{
     	\includegraphics[width=0.18\textwidth]{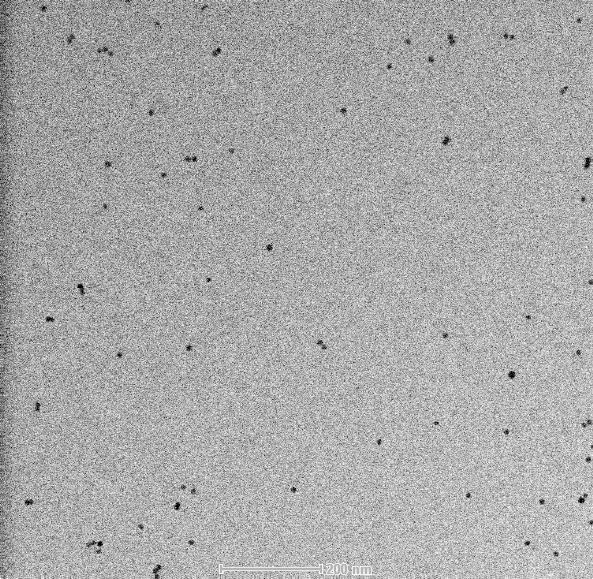}
}
  \subfigure[image 5]{    
	\includegraphics[width=0.18\textwidth]{images/Image112.jpeg}	
}   
	  \subfigure[image 6]{    
		\includegraphics[width=0.18\textwidth]{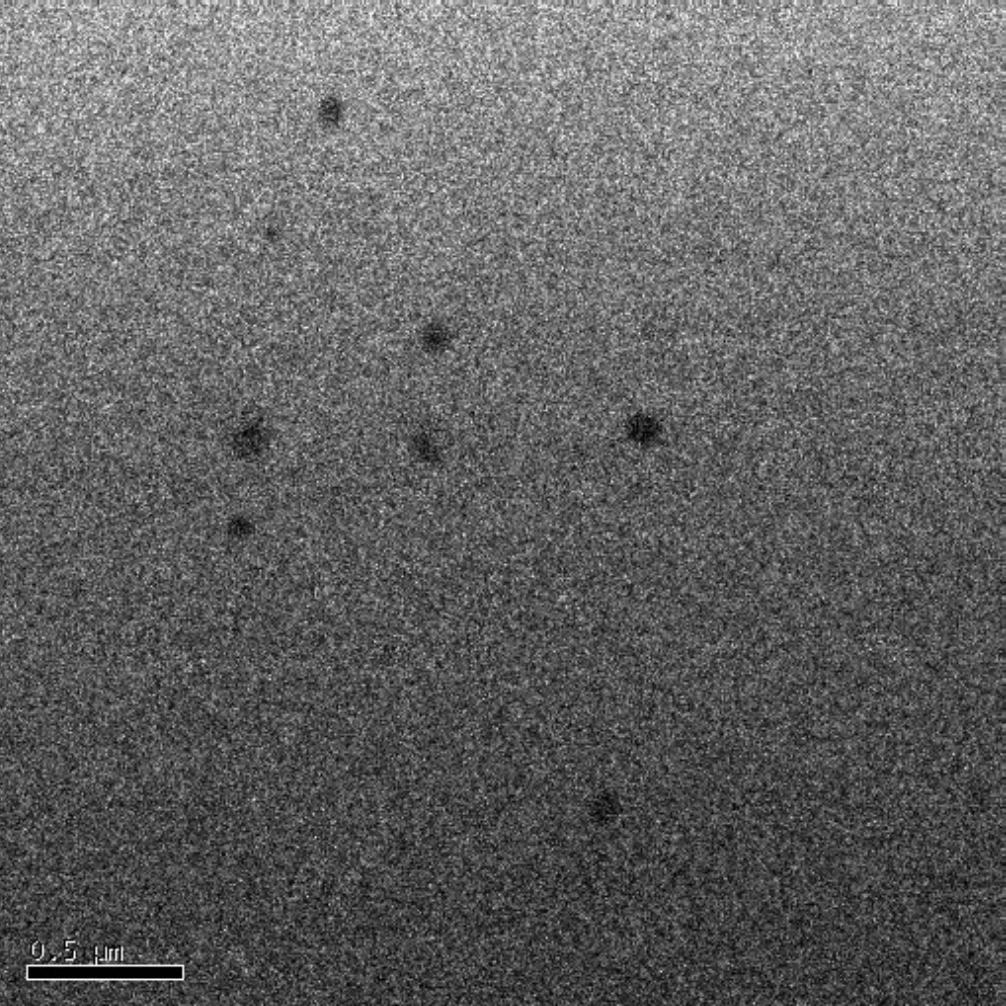}	
	}   
\subfigure[image 7]{
     	\includegraphics[width=0.18\textwidth]{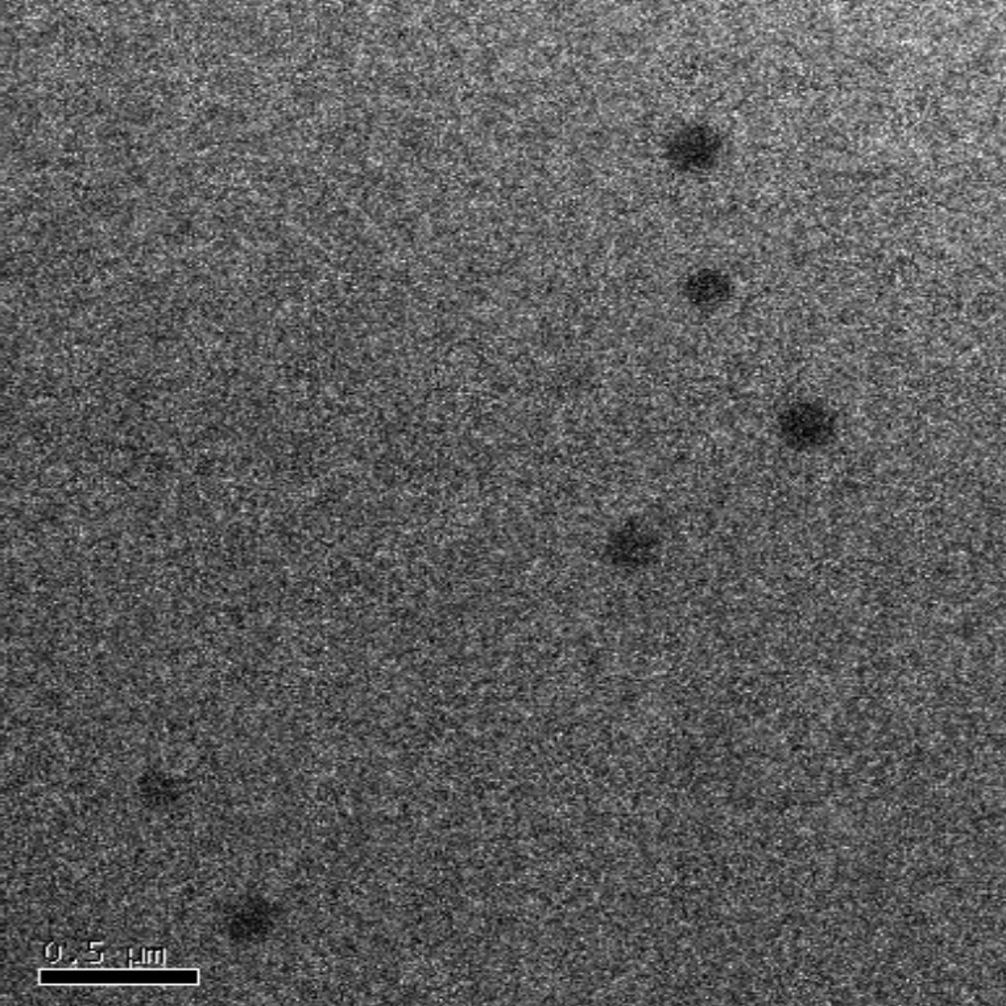}
}
    \subfigure[image 8]{    
	\includegraphics[width=0.18\textwidth]{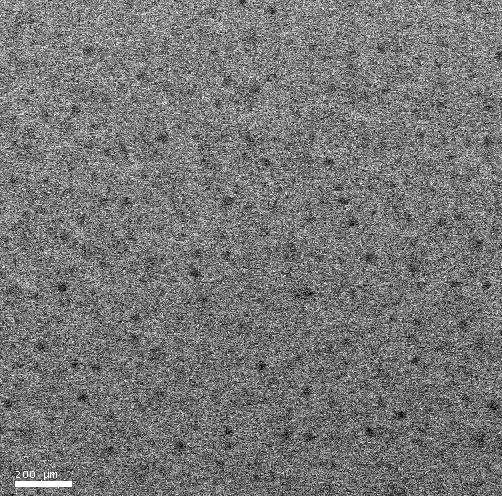}
	}
\caption{\texttt{NANOPARTICLE} Dataset}
\label{fig_dataset3}
\end{figure}

We evaluated our approach using three benchmark datasets that consist of 26 test images and the corresponding ground truth binary images. The first benchmark dataset is \texttt{DIBCO 2011}, which was provided as a part of the 2011 ICDAR document image binarization contest \cite{pratikakis2013icdar}. As shown in Figure \ref{fig_dataset1}, the dataset consists of eight optical character recognition document images and their groundtruth binary images. The second benchmark dataset is \texttt{H-DIBCO 2016}, which was provided as a part of the 2016 ICFHR handwritten document image binarization contest \cite{pratikakis2016icfhr2016}, and it comes with ten scan images of handwritten documents as shown in Figure \ref{fig_dataset2}. The third dataset is \texttt{NANOPARTICLE}, which contains eight electron microscope images and corresponding ground truth binary images as shown in Figure \ref{fig_dataset3}. The microscope images were experimentally produced using a high-resolution transmission electron microscope by our collaborators at Pacific Northwest National Lab, and each of them contains tens to hundreds of nanoparticles over an uneven and noisy background. The ground truth binary images for the first two datasets are given as parts of the datasets, and those for the last dataset were manually generated in our lab.\\

We used the benchmark datasets to perform two kinds of evaluation. The first evaluation is to test the image binarization performance of our approach, with comparison to nine other local adaptive thresholding methods. The second evaluation is to show how the proposed approach improves the accuracy of morphological image segmentation methods in terms of segmenting overlapping foreground objects, because we believe that an improved binarization method can significantly improve the accuracy of morphological image segmentation methods that use the binary silhouette of foregrounds as an input. The outcomes of the first evaluation is summarized in Section \ref{sec:bin_outcome}, and the outcomes of the second evaluation is summarized and discussed in Section \ref{sec:seg_outcome}. 

\subsection{Image Binarization Outcomes} \label{sec:bin_outcome}
 We compared the binarization performance of our proposed approach with nine state-of-the-art image binarization methods including NIBLACK \cite{niblack1985introduction}, BERNSE \cite{bernse1986dynamic}, GATOS\cite{gatos2006adaptive}, BRADLEY \cite{bradley2007adaptive}, SAUV \cite{sauvola2000adaptive}, PHAN \cite{phansalkar2011adaptive}, LU \cite{lu2010document}, SU \cite{su2013robust}, and HOWE \cite{Howe2013}; the HOWE was the base of the first place winner of the 2016 ICFHR handwritten document image binarization contest \cite{pratikakis2016icfhr2016}. The performance metrics that were used for the document binarization contest are applied in this paper, including the F-measure (FM), pseudo F-measure (PFM), peak signal-to-noise ratio (PSNR),  distance reciprocal distortion metric (DRD) and misclassification penalty metric (MPM). The FM is based on the pixel-wise binarization recall and precision, 
\begin{equation*}
FM = \frac{2 \times RC \times PR}{RC+PR},
\end{equation*}
where $RC$ and $PR$ are the binarization recall and precision respectively. The PFM uses the pseudo binarization recall and precision, which consider distance weights of pixels to the nearest contours of foregrounds in computing the recall and precision. The PSNR is $10\log(C^2 / MSE)$, where $MSE$ and $C$ refer the mean square error and the average intensity level difference between foregrounds and backgrounds. The DRD has been used to measure the visual distortion of an estimated binary image from its ground truth counterpart; its complex formula is given in literature \cite{lu2004distance}. The MPM is defined as 
\begin{equation*}
MPM = \frac{\sum_{i=1}^{N_{FN}} d^{i}_{FN} + \sum_{j=1}^{N_{FP}} d^{j}_{FP}}{2D},
\end{equation*}
where $D$ is a scaling constant, $N_{FN}$ is the number of false negatives, $N_{FP}$ is the number of false positives respectively, $d^{i}_{FN}$ is the distance from the $i$th false negative to the nearest foreground contour pixel in the ground truth, and $d^{i}_{FP}$ is the distance from the $i$th false positive to the nearest foreground contour pixel in the ground truth. Higher FM, PFM and PSNR are better, while lower $DRD$ and $MPM$ are better. \\

\begin{figure}
\centering
\includegraphics[width=0.9\textwidth]{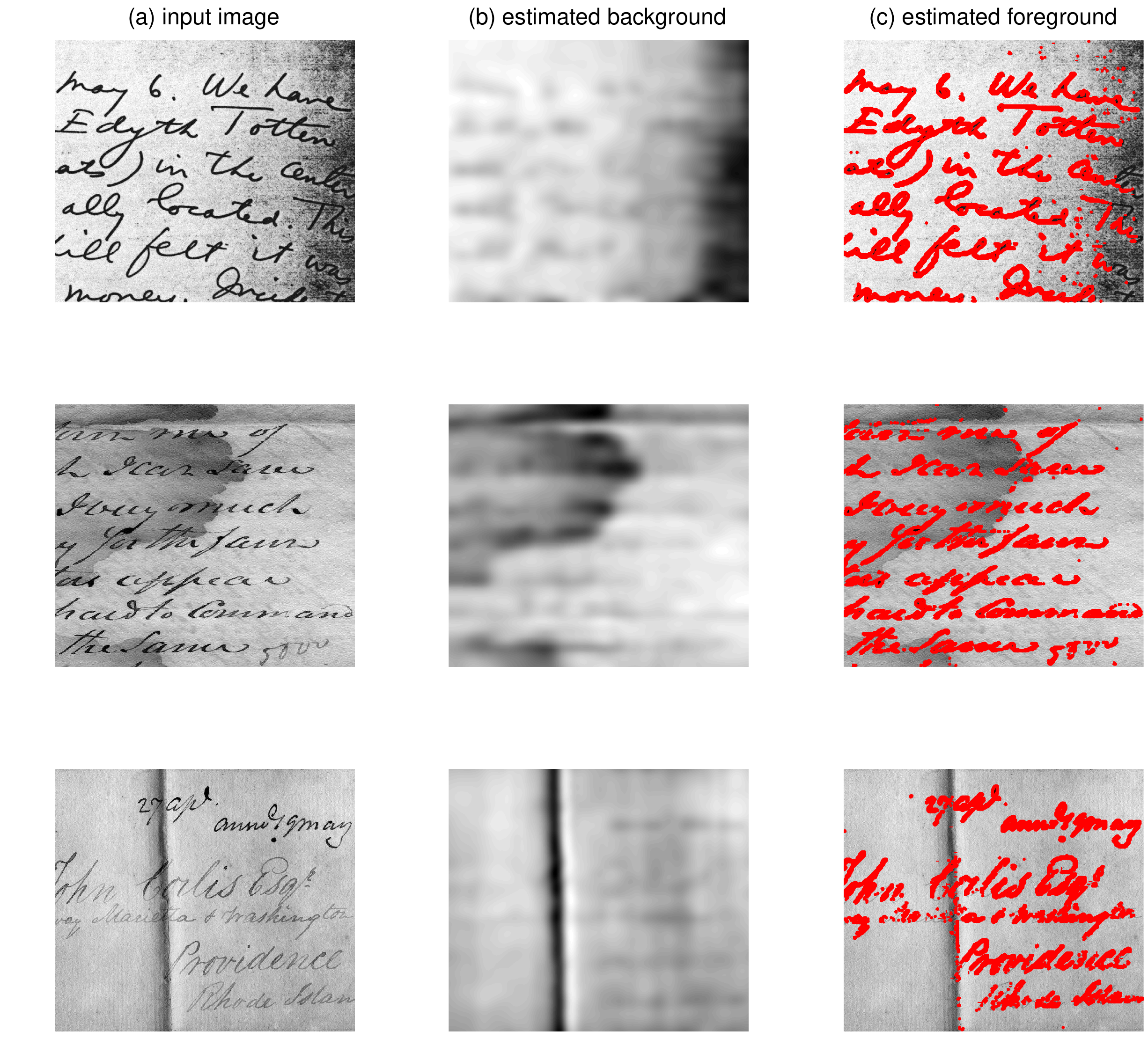}
\caption{Results of the proposed approach for the \texttt{DIBCO 2011} dataset}
\label{outcome_fig1}
\vspace{14pt}
\begin{tabular}{|l|c|c|c|c|c|}
\hline
Dataset: \texttt{DIBCO 2011} 	& FM & Pseudo FM & PSNR & DRD & MPM \\
\hline
NIBLCK \cite{niblack1985introduction} &  34.6304  & 35.1075 &  5.8505 & 91.2962 & 0.1802 \\
BERNSE \cite{bernse1986dynamic}       &  48.3548  & 51.5734 &  8.8261 & 81.1017 & 0.1352 \\
GATOS \cite{gatos2006adaptive}        &  75.3618  & 85.2602 & 15.6972 & 6.0472 & 0.0013 \\ 
BRADLEY \cite{bradley2007adaptive}    &  78.1436  & 82.0379 & 15.1789 & 11.3524 & 0.0207 \\
SAUV\cite{sauvola2000adaptive}		&  83.6913  & 88.5199 & 16.9442 & 5.1378 & 0.0052  \\
PHAN  \cite{phansalkar2011adaptive}          &  81.1714  & 87.5590 & 16.3234 & 6.0770 & 0.0069  \\
LU    \cite{lu2010document}        & 85.2025  & 86.5061  & 17.1189 &   5.4913  &  0.0085\\
SU \cite{su2013robust}             &  88.9265  & 90.9494 & 18.7875 & 3.9674 & 0.0054  \\
HOWE  \cite{Howe2013}           &  89.2447  & 90.3298 & 20.0755 & 2.8861 & 0.0007  \\
\hline
PROPOSED        &  88.2467  & 89.6248 & 17.8437 & 4.4398 & 0.0041  \\
\hline
\end{tabular}
\captionof{table}{Five performance metrics of nine state-of-the-art image binarization methods and our proposed approach for the \texttt{DIBCO 2011} dataset.} 
\label{tbl:dibco11}
\end{figure}

\begin{figure}
\centering
\includegraphics[width=0.9\textwidth]{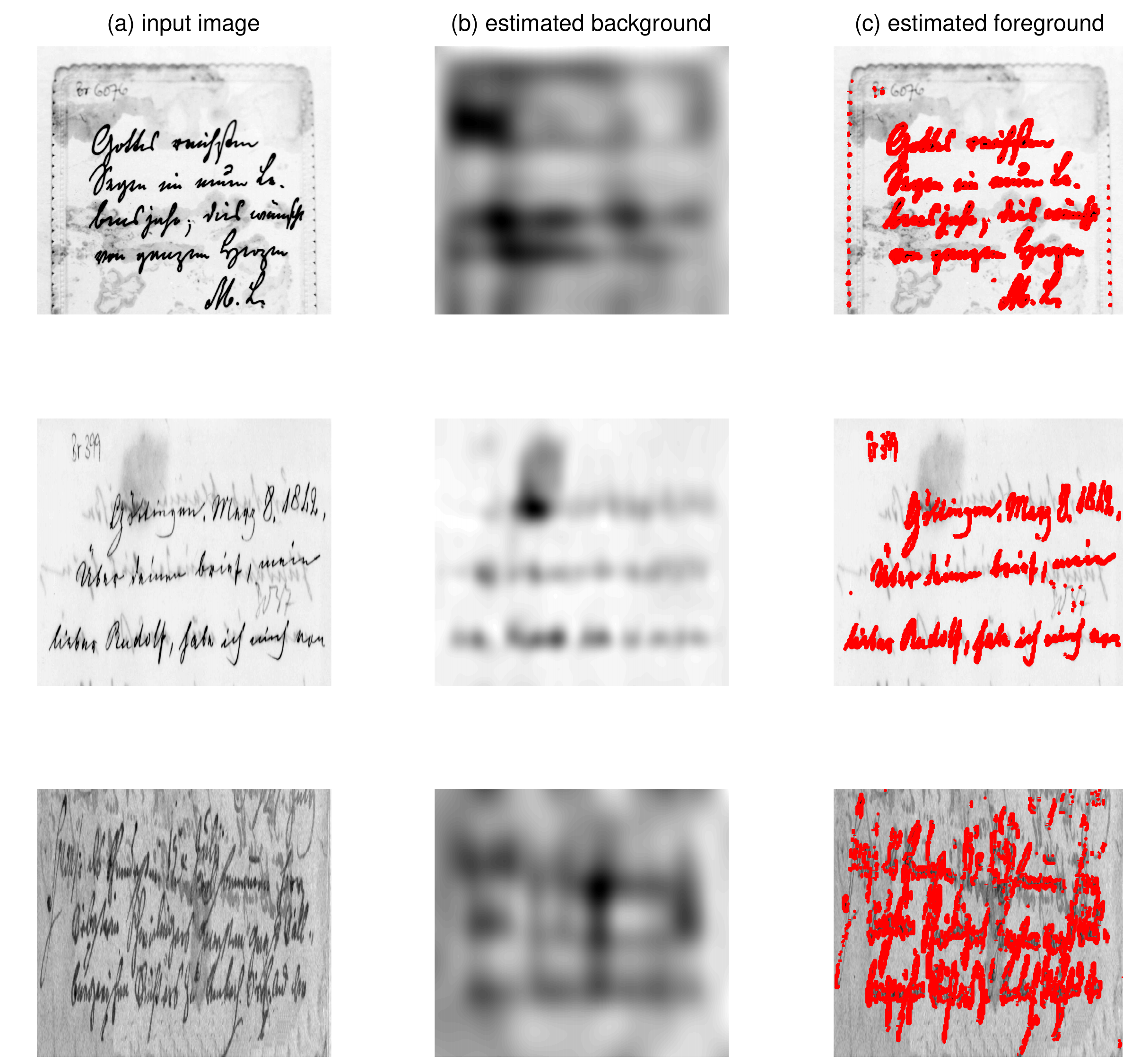}
\caption{Results of the proposed approach for the \texttt{H-DIBCO 2016} dataset}
\label{outcome_fig2}
\vspace{14pt}
\centering
\begin{tabular}{|l|c|c|c|c|c|}
\hline
Dataset: \texttt{H-DIBCO 2016} 	& FM & Pseudo FM & PSNR & DRD & MPM \\
\hline
NIBLCK \cite{niblack1985introduction} &  38.8471   & 39.2807  & 6.4298   & 118.1658 & 0.1607 \\
BERNSE \cite{bernse1986dynamic}       &  61.3108   & 65.6364  & 11.8619  & 23.3392 &  0.0312 \\
GATOS \cite{gatos2006adaptive}        &  72.6761   & 81.3539  & 14.8873  & 24.7338 &  0.0512\\ 
BRADLEY \cite{bradley2007adaptive}    &  82.1053   & 86.1670  & 15.8005  &  9.0341 &  0.0100 \\
SAUV\cite{sauvola2000adaptive}		&  86.1422   & 90.6095  & 17.8828  &  4.8947 &  0.0025   \\
PHAN  \cite{phansalkar2011adaptive} &  81.6341   & 86.6807  & 17.3355  &  6.2056 &  0.0032   \\
LU    \cite{lu2010document}        & 86.8274 &  90.3314 &  17.9318 &   4.9821  &  0.0058\\
SU \cite{su2013robust}             & 85.1194   & 90.2589  & 17.5614  &  5.5390 &  0.0047  \\
HOWE  \cite{Howe2013}           & 88.1152   & 92.7556  & 18.3030  &  4.3817 &  0.0036  \\
\hline
PROPOSED        &  87.2611   & 90.4800  & 17.5358  &  3.6832 &  0.0019  \\
\hline
\end{tabular}
\captionof{table}{Five performance metrics of nine state-of-the-art image binarization methods and our proposed approach for the \texttt{H-DIBCO 2016} dataset.} 
\label{tbl:dibco16}
\end{figure}

Table \ref{tbl:dibco11} presents the numbers of the performance metrics for the first dataset, and Table \ref{tbl:dibco16} presents those for the second dataset. The numbers in the tables are the performance metrics averaged over all test images in each dataset. The performance metrics of our proposed approach are very comparable to those of the two top performers for the first dataset and the second dataset. The first two datasets contain little noises but complex background patterns due to document degradation and imperfectly erased handwritings. Our approach has shown competitive performance in handling such complexities. In particularly, the proposed approach performed better than the existing approaches for image 1 and image 4 in the first dataset and for images 4, 8, and 10 in the second dataset. Those images have more complicated background variations than the other documents images. Figures \ref{outcome_fig1} and \ref{outcome_fig2} show illustrative outcomes of our approach for those images. In the figures, we can see that the backgrounds estimated by our approach successfully captured the complex patterns of document image backgrounds, so the foreground estimates were not much affected by the image backgrounds.\\

\begin{figure}
\centering
\includegraphics[width=0.9\textwidth]{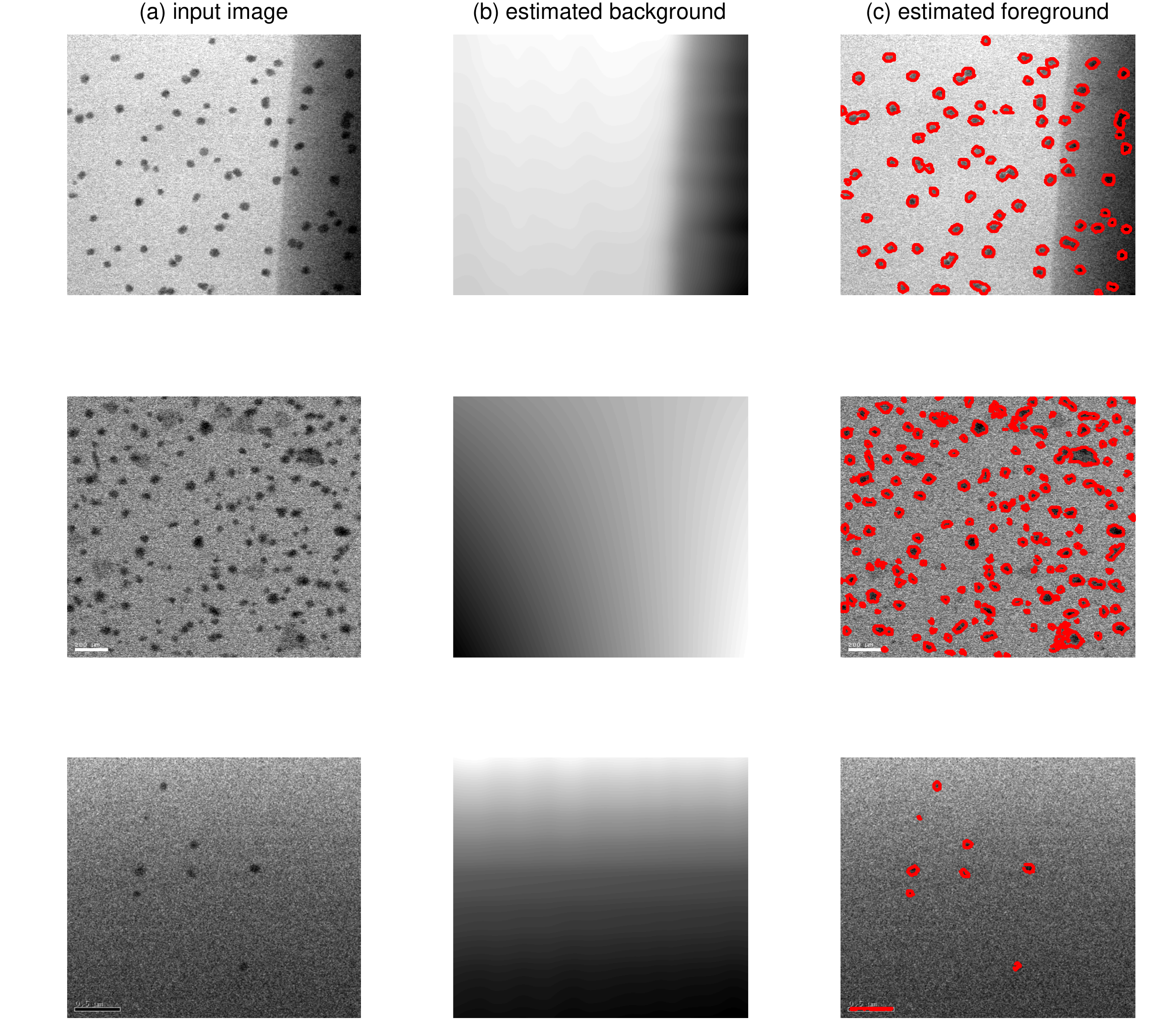}
\caption{Results of the proposed approach for the \texttt{NANOPARTICLE} dataset}
\label{outcome_fig3}
\vspace{14pt}
\begin{tabular}{|l|c|c|c|c|c|}
\hline
Dataset: \texttt{NANOPARTICLE} 	& FM & Pseudo FM & PSNR & DRD & MPM \\
\hline
NIBLCK \cite{niblack1985introduction} &  30.5725   & 31.5958  & 4.6986   & 965.9741  &  0.1912 \\
BERNSE \cite{bernse1986dynamic}       &  29.8622   & 30.5462  & 4.0922   & 1117.9973 &  0.2325 \\
GATOS \cite{gatos2006adaptive}        &  39.0873   & 46.3769  & 10.2806  & 205.7941  &  0.0728\\ 
BRADLEY \cite{bradley2007adaptive}    &  35.9182   & 37.7035  & 6.2595   & 760.9104  &  0.1274 \\
SAUV\cite{sauvola2000adaptive}		&  40.8041   & 42.7942  & 8.1357   & 380.8603  &  0.0772   \\
PHAN  \cite{phansalkar2011adaptive} &  39.6126   & 42.3593  & 7.8169   & 472.2628  &  0.0802   \\
LU \cite{lu2010document}           & 24.6024& 	24.7608   &	3.0811 & 920.42172051914 &	0.3463 \\
SU \cite{su2013robust}             & 25.0829   & 23.8462  & 5.5022   & 719.4512  &  0.1389  \\
HOWE  \cite{Howe2013}           & 37.5394   & 37.9925  & 11.1987  & 229.1684  &  0.0354  \\
\hline
PROPOSED        &  80.7743   & 87.7246  & 17.6784  & 10.8957   &  0.0036   \\
\hline
\end{tabular}
\captionof{table}{Five performance metrics of nine state-of-the-art image binarization methods and our proposed approach for the \texttt{NANOPARTICLE} dataset.} 
\label{tbl:nano}
\end{figure}

On the other hand, our proposed approach outperformed the nine state-of-the-art binarization methods significantly for the last dataset. Table \ref{tbl:nano} summarizes the five performance metrics for the last dataset. The major difference of the last dataset from the previous two datasets is that the last dataset has significantly higher background noises and larger foreground sizes. When a noise level is very high (i.e., the signal-to-noise ratio of an input image is low), local image contrasts are significantly affected by image noises, which causes the methods based on a local image contrast map (such as \cite{su2013robust}) less competitive for the last dataset. When foreground sizes are large and noises are severe, estimating the image background accurately is quite challenging. This is why the background subtraction method such as \cite{lu2010document} \cite{gatos2006adaptive} did not work well for the last dataset. Under the circumstances, our proposed approach is still able to capture the background very robustly. The background estimates of our approach are presented in Figure \ref{outcome_fig3}, which shows the robustness of our background estimator. 

\subsection{Effects on Image Segmentation} \label{sec:seg_outcome}
Our approach performed well in image binarization for all 26 test images. The great image binarization outcomes of the proposed approach can be used in a morphological image segmentation method to improve the overall accuracy of identifying individual foreground objects under overlaps. This section shows how our approach can improve the existing morphological image segmentation methods. \\

We use the \texttt{NANOPARTICLE} dataset for this study, which contains eight microscope images of overlapping nanoparticles; the other two datasets do not have any foreground overlap issues, so those were not included in this study. Table \ref{tbl:nano_data_desc} summarizes the characteristics of the eight images in terms of the signal-to-noise ratio (SNR), the background variation and the foreground density; the definitions of the characteristics are described in the table caption. Images having higher foreground densities contain more foreground overlaps, e.g., Img 1, Img 2 and Img 5. Three morphological image segmentation methods are considered in this testing, including ultimate erosion for convex sets (UECS) \cite{park2013segmentation}, bounded erosion with fast radial symmetry (BE-FRS) \cite{zafari2015segmentation} and morphological multiscale method (MSD) \cite{schmitt2009morphological}, which are specialized for segmenting overlapping objects in microscope images. We counted the number of falsely identified foreground objects (false positives) and the number of unidentified identified foreground objects (false negatives) for the three methods when their built-in image binarization are applied as well as when our approach replaced the built-in binarization. \\

\begin{figure}
\centering
\includegraphics[width=\textwidth]{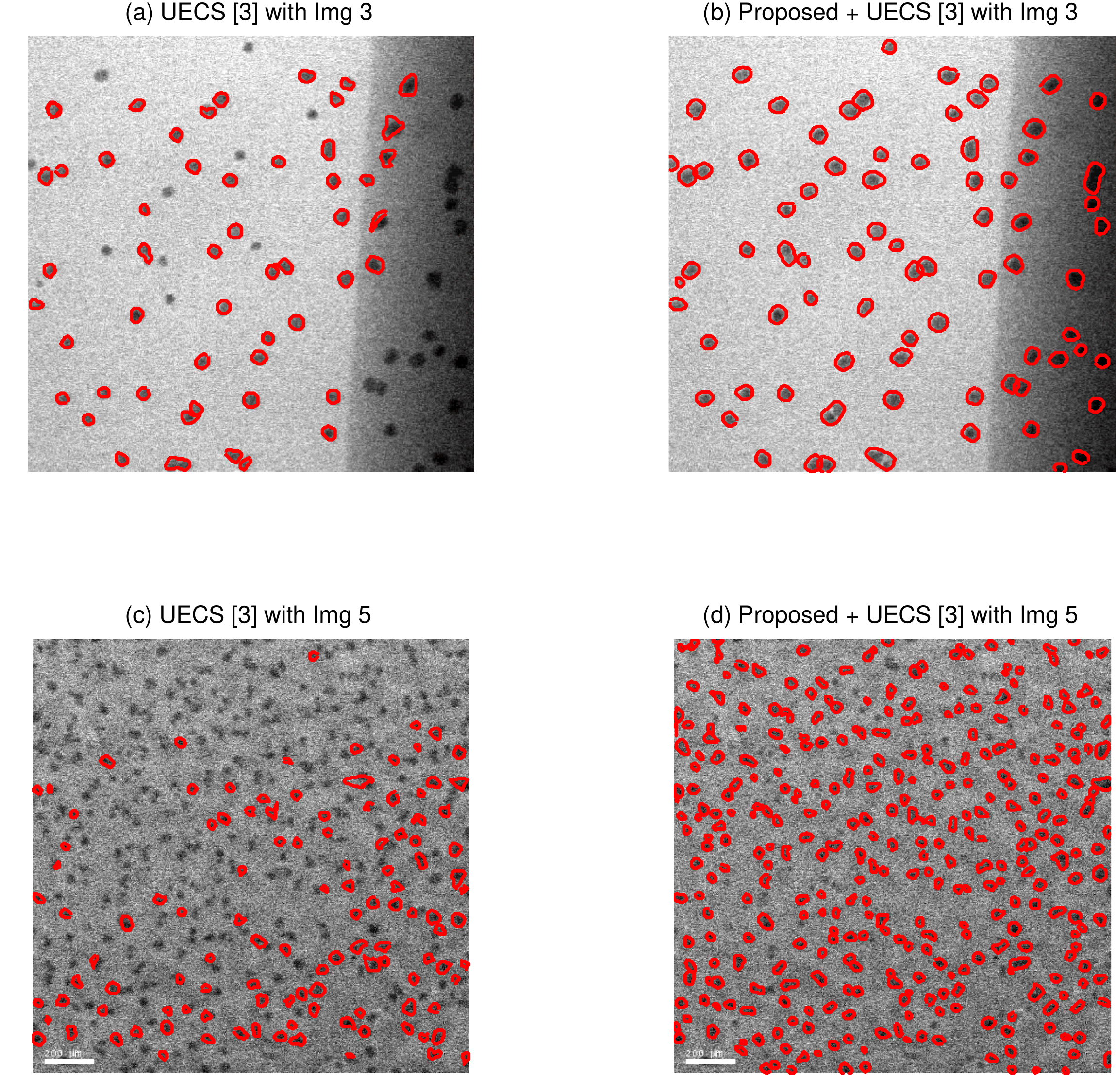}
\caption{Comparison of UECS \cite{park2013segmentation} with UECS + Proposed Approach. With significant background variation, the original UECS missed a number of foreground objects. If the binarization outcome of the proposed approach is used in the UECS, the foreground detection can be significantly improved.}
\label{outcome_fig4}
\end{figure}

Table \ref{tbl:seg_outcome} summarizes the number of false positives and false negatives for the eight test images. The morphological image segmentation methods produced significantly many false negatives and false positives in particular for the co-existence of background variations and high noise levels such as Img 3 and Img 5. This is mainly because the built-in image binarization algorithms in the morphological image segmentation methods worked poorly for the test images. Using the proposed binarization outcome as input to the morphological image segmentation methods significantly reduced the numbers of false positives and false negatives. Figure \ref{outcome_fig4} illustrates the image segmentation outcomes of the UECS \cite{park2013segmentation} and the proposed approach combined with the UECS for Img 3 and Img 5. This illustration clearly shows how the proposed approach improved the existing morphological image segmentation methods for complex image segmentation works. 

\begin{table}
\centering
\begin{tabular}{|l|r|r|r|r|}
\hline
      & \# of Foregrounds & SNR($\sigma_{fg}^2/\sigma^2_{noise}$) & $\sigma_{bg}^2$ & Foreground Density \\
      \hline
Img 1 & 86 & 31.22 & 112.56 &  26.80\% \\ 
Img 2 & 162& 1.53 & 1.15 & 11.28\% \\  
Img 3 & 83 & 2.13 & 2719.80 & 5.88\% \\ 
Img 4 & 69 & 1.51 & 81.02 & 0.61\% \\ 
Img 5 & 351& 1.41 & 91.05 & 17.74\% \\ 
Img 6 & 8  & 1.31  &  787.01 &  0.72\% \\ 
Img 7 & 6  & 1.36  & 174.69 & 1.14\% \\ 
Img 8 & 140& 1.14 &  56.24 & 2.80\% \\  
\hline
\end{tabular}
\caption{Characteristics of test microscope images. The $\sigma^2_{fg}$ is the variance of image intensities, $\sigma^2_{noise}$ is the variance of noises, and $\sigma_{bg}^2$ is the variance of background intensities. The foreground density is the fraction of the number of foreground pixels in each image.}
\label{tbl:nano_data_desc}
\end{table}

\begin{table}
\centering
\begin{tabular}{|l|r|r|r|r|r|r|r|r|r|r|r|r|}
\hline
             & \multicolumn{2}{c}{UECS\cite{park2013segmentation}} & \multicolumn{2}{|c}{Proposed} & \multicolumn{2}{|c}{FRS\cite{zafari2015segmentation}}    & \multicolumn{2}{|c}{Proposed}    & \multicolumn{2}{|c}{MSD\cite{schmitt2009morphological}}        & \multicolumn{2}{|c|}{Proposed} \\
             & \multicolumn{2}{c}{}                               & \multicolumn{2}{|c}{+ UECS\cite{park2013segmentation}}   & \multicolumn{2}{|c}{}   &  \multicolumn{2}{|c}{+ FRS\cite{zafari2015segmentation}}      &  \multicolumn{2}{|c}{}  & \multicolumn{2}{|c|}{+ MSD\cite{schmitt2009morphological}} \\
             \cline{2-13}
             & FN & FP & FN & FP & FN & FP & FN & FP & FN & FP & FN & FP \\
             \hline
Img 1 & 3  & 0  & 1 & 0           & 3& 1   & 1& 0         & 3 & 0 	& 6 & 0\\ 
Img 2 & 21 & 5  & 2& 18         & 12& 20  & 3 & 17         & 0& 499	& 8 & 11\\ 
Img 3 & 32 & 2  & 10& 2         & 17& 30  & 9& 0         & 30& 1 	& 10& 2\\ 
Img 4 & 69 & 0 &  23& 4 & 52& 5   & 39& 9         & 65& 838 	& 15 & 7\\ 
Img 5 & 183 & 7  & 18& 0         & 67& 17  & 26& 0         & 201 & 796    & 18& 0\\ 
Img 6 & 2 & 0   & 0& 0           & 1& 603  & 0& 0         & 7 & 19 	& 0 & 0\\ 
Img 7 & 1 & 2 &   0& 1          & 2& 1419& 0& 1         & 5& 707      & 0& 1\\ 
Img 8 & 138& 0 & 32& 0         & 76& 24  & 60& 0         & 26& 704 	& 32& 0\\ 
\hline
\end{tabular}
\caption{Performance of the existing morphological image segmentation methods combined with the proposed image binarization. FN = the number of unidentified foreground objects. FP = the number of falsely identified foregrounds.}
\label{tbl:seg_outcome}
\end{table}

\section{Conclusion} \label{con}
We presented a new approach that solves an image binarization problem under significant background variations and noises. The approach basically estimates the background intensity variation of an input image and subtracts the estimate from the input to achieve a flat-background image, which is thresholded by a global thresholding approach to get the final binary outcome. For robust estimation of a background intensity variation, we proposed a robust regression approach to recover a smooth intensity surface of an image background which is buried under foreground intensities and noise intensities. A global threshold selector was proposed, based on a model selection criterion. With the improved background estimator and threshold selector, the proposed approach has shown great binarization performance quite uniformly for all of our 26 benchmark images including 18 document images and eight microscopy images. We also showed how the improved binarization performance can be used for complex image segmentation works of segmenting overlapping foregrounds under uneven background and heavy noises. We believe that the proposed approach has great values in robust image binarization and image segmentation. 

\section*{Acknowledgment}
The authors would like to acknowledge support for this
project. This work is partially supported by NSF 1334012,
AFOSR FA9550-13-1-0075, AFOSR FA9550-16-1-0110, and
FSU PG 036656.



\section*{References}
\bibliographystyle{elsarticle-num} 
\bibliography{nanoimage}
\end{document}